\documentclass[sigconf]{acmart}
\AtBeginDocument{%
  }

\usepackage{thmtools}
\usepackage{enumitem}
\usepackage{amsmath}
\usepackage{pifont}
\usepackage{subcaption}
\usepackage{cleveref}
\usepackage[ruled,linesnumbered]{algorithm2e}
\usepackage{multirow}
\usepackage{float}

\AtEndPreamble{%
 \theoremstyle{acmplain}
 }

\AtEndPreamble{%
 \theoremstyle{acmplain}
 }

\newcommand{\name}{\textbf{TFMLinker}}
\newcommand{\Context}{Prototype-Augmented Local-Global Context}
\newcommand{\Encoder}{Universal Topology-Aware Link Encoder}
\newcommand{\context}{prototype-augmented local-global context}
\newcommand{\encoder}{universal topology-aware link encoder}

\newcommand{\best}[1]{\textbf{#1}} %
\newcommand{\secbest}[1]{\underline{#1}} %

\setcopyright{acmlicensed}
\copyrightyear{2026}
\acmYear{2026}
\acmDOI{XXXXXXX.XXXXXXX}
\acmConference[KDD '26]{Proceedings of the 32nd ACM SIGKDD Conference on Knowledge Discovery and Data Mining}{August 09--13, 2026}{Jeju Island, Republic of Korea}

\acmSubmissionID{2772}

\begin{document}

\title{TFMLinker: Universal Link Predictor by Graph In-Context Learning with Tabular Foundation Models}

\author{Tianyin Liao}
\affiliation{
  \institution{Nankai University}
  \city{Binhai New Area}
  \state{Tianjin}
  \country{China}
}
\email{1120230329@mail.nankai.edu.cn}

\author{Chunyu Hu}
\affiliation{
  \institution{Nankai University}
  \city{Binhai New Area}
  \state{Tianjin}
  \country{China}
}
\email{huchunyu@mail.nankai.edu.cn}

\author{Yicheng Sui}
\affiliation{
  \institution{Nankai University}
  \city{Binhai New Area}
  \state{Tianjin}
  \country{China}
}
\email{suiyicheng@nankai.edu.cn}

\author{Xingxuan Zhang}
\affiliation{
  \institution{Tsinghua University}
  \city{Haidian District}
  \state{Beijing}
  \country{China}
}
\email{xingxuanzhang@hotmail.com}

\author{Peng Cui}
\affiliation{
  \institution{Tsinghua University}
  \city{Haidian District}
  \state{Beijing}
  \country{China}
}
\email{cuip@tsinghua.edu.cn}

\author{Jianxin Li}
\affiliation{
  \institution{Beihang University}
  \city{Haidian District}
  \state{Beijing}
  \country{China}
}
\email{lijx@buaa.edu.cn}

\author{Ziwei Zhang}
\authornote{Corresponding author.}
\affiliation{
  \institution{Beihang University}
  \city{Haidian District}
  \state{Beijing}
  \country{China}
}
\email{zwzhang@buaa.edu.cn}

\renewcommand{\shortauthors}{Tianyin Liao et al.}

\begin{abstract}
Link prediction is a fundamental task in graph machine learning with widespread applications such as recommendation systems, drug discovery, knowledge graphs, etc. In the foundation model era, how to develop universal link prediction methods across datasets and domains becomes a key problem, with some initial attempts adopting Graph Foundation Models (GFMs) utilizing Graph Neural Networks (GNNs) and Large Language Models (LLMs). However, the existing methods face notable limitations, including limited pre-training scale or heavy reliance on textual information. Motivated by the success of tabular foundation models (TFMs) in achieving universal prediction across diverse tabular datasets, we explore an alternative approach by TFMs, which are pre-trained on diverse synthetic datasets sampled from structural causal models and support strong in-context learning independent of textual attributes. Nevertheless, adapting TFMs for link prediction faces severe technical challenges such as how to obtain the necessary context and capture link-centric topological information. To solve these challenges, we propose \name~ (\underline{T}abular \underline{F}oundation \underline{M}odel for \underline{Link} Predict\underline{or}), aiming to leverage the in-context learning capabilities of TFMs to perform link prediction across diverse graphs without requiring dataset-specific fine-tuning. Specifically, we first develop a \context~module to construct context that captures both graph-specific and cross-graph transferable patterns. Next, we design a \encoder~to capture link-centric topological information and generate link representations as inputs for the TFM. Finally, we employ the TFM to predict link existence through in-context learning. Experiments on 6 graph benchmarks across diverse domains demonstrate the superiority of our method over state-of-the-art baselines without requiring dataset-specific finetuning.
\end{abstract}



\maketitle

\section{Introduction}\label{sec:intro}
Link prediction is a fundamental task in graph machine learning with broad applications across social networks, biology, recommendation systems, and beyond~\cite{kovacs2019network,daud2020applications}. The goal is to infer missing or potential links within a graph from observed ones. Traditional link prediction methods typically rely on hand-crafted heuristics to model pairwise node relationships, such as common neighbors or shortest path distances. While graph neural networks (GNNs) have recently advanced link prediction performance~\cite{chamberlaingraph,yun2021neo,wangneural,shomer2024lpformer}, existing methods still face key challenges, including limited supervision due to sparse connectivity, sensitivity to initialization, and poor generalization under distribution shifts~\cite{song2025scalable,li2023evaluating}, which collectively constrain their applicability across diverse datasets and domains.

In the foundation model era, addressing these challenges to develop universal link prediction methods across datasets and domains becomes a key problem. Graph Foundation Models (GFMs), envisioned as models pre-trained on extensive graph data, have been proposed as promising solutions adaptable to diverse link prediction scenarios. Architecturally, GFMs can be classified into three main types~\cite{liu2025graph}: GNN-based~\cite{song2025scalable,dong2025universal}, LLM-based~\cite{chen2024exploring}, and GNN+LLM-based models~\cite{liuone}. These architectures leverage different underlying mechanisms to enhance their adaptability and generalization across diverse link prediction tasks. For example, GNN-based models can utilize graph-structured inductive biases to capture topological patterns, LLM-based models incorporate semantic understanding through natural language processing, and GNN+LLM-based models synergize structural and semantic information for more generalized representations. %

However, despite their initial progress, these architectures all face severe limitations. For instance, GNN-based GFMs typically lack pre-training on sufficiently diverse and comprehensive datasets, which limits their ability to generalize across different domains and constrains their performance in tasks requiring broad contextual understanding~\cite{liu2025graph}. Meanwhile, both LLM-based and GNN+LLM-based GFMs often rely heavily on textual information, leading to inconsistent data priors when applied to graphs that lack sufficient textual attributes~\cite{eremeev2025turning,eremeev2025graphpfn}.

To overcome these limitations, and inspired by the success of tabular foundation models (TFMs) in achieving universal prediction across diverse tabular datasets, we propose exploring universal link prediction methods based on leading TFMs~\cite{hollmanntabpfn,hollmann2025accurate,zhang2025limix}. These methods are pre-trained on diverse synthetic datasets sampled from structural causal models and can make predictions in an in-context learning setting. Recent works, such as TabPFNv2~\cite{hollmann2025accurate}, LimiX~\cite{zhang2025limix}, and GraphPFN~\cite{eremeev2025graphpfn}, have already demonstrated strong generalization capabilities independent of textual attributes, performing well on both tabular prediction and graph node classification tasks. Nevertheless, directly applying such TFMs to the link prediction problem faces significant challenges, primarily arising from the following aspects.

First, how to obtain the context necessary for such TFMs remains unexplored in link prediction tasks. A straightforward approach is to randomly downsample edges from the target graph to gain context. However, relying solely on the target graph only allows the model to learn patterns within a single task but neglects capturing the common and transferable patterns that exist across diverse tasks and graphs~\cite{wumoemeta}.

Second, how to enable TFMs to capture link-centric topological information remains unexplored. TabPFN and LimiX have been proven to be a context-learning model capable of handling data heterogeneity~\cite{ye2025closer}, showing potential in various domains. However, given their nature as tabular foundation models pre-trained on synthetic tabular data, they cannot inherently capture link-centric topological information or derive link representations from datasets rich in structural diversity.

To address these challenges, we propose \name~ (\underline{T}abular \underline{F}oundation \underline{M}odel for \underline{Link} Predict\underline{or}), aiming to leverage the in-context learning capabilities of TFMs to perform link prediction across diverse graphs on-the-fly, without requiring explicit model fine-tuning tailored to specific downstream graph datasets. Specifically, \name~achieves this goal by tailoring three modules for universal link prediction. First, we design a \context~module to construct context through random subgraph downsampling and learnable prototypes. This dual mechanism empowers our method to capture two distinct types of pattern: specific patterns within individual graphs using downsampling, and transferable patterns across diverse graphs using prototypes, respectively. Next, we introduce a \encoder~to capture link-centric topological information by multiple link encoders. In this way, we utilize the strengths of both heuristic link-centric methods and the pre-trained GNN to generate link representations, with each embedding vector forming a row of a pseudo table and used as inputs to the TFM, thereby providing it with rich topological information. Finally, we employ the TFM to predict link existence through in-context learning. We conduct extensive experiments on 6 graph benchmarks across various domains for link prediction. The results demonstrate that \name~consistently and significantly outperforms various state-of-the-art baselines without requiring fine-tuning on target graphs. In-depth analyses further demonstrate the designs of our proposed method.

Our contributions are summarized as follows:
\begin{itemize}[leftmargin=0.5cm]
    \item We propose \name, a novel link prediction approach across datasets and domains leveraging the in-context learning abilities of TFMs. To the best of our knowledge, this work is the first exploration of adapting TFMs to the link prediction task.
    \item We develop a \context~module to construct context via random subgraph downsampling and learnable prototypes, enabling the capture of both graph-specific and cross-graph transferable patterns.
    \item We further design a \encoder~module to capture link-centric topological information by multiple link encoders.%
    \item Extensive empirical results demonstrate \name~outperforms state-of-the-art methods across datasets from diverse graph domains without requiring dataset-specific fine-tuning.
\end{itemize}

The rest of the paper is organized as follows. In Section~\ref{sec:form}, we provide the necessary preliminary knowledge and the problem formulation. In Section~\ref{sec:method}, we introduce our proposed method in detail. In Section~\ref{sec:exp}, we present a comprehensive evaluation, detailing our experimental setup, reporting comparative results, and offering an in-depth analysis of the findings. We review related works in Section~\ref{sec:related} and conclude the paper in Section~\ref{sec:con}.

\section{Preliminary and Problem Formulation}\label{sec:form}

\paragraph{Link prediction.} Consider a graph $\mathcal{G}=(\mathcal{V},\mathcal{E}^o)$, where $\mathcal{V}=\{v_1,v_2,\cdots,v_n\}$ is the node set of cardinality $n$, and $\mathcal{E}^o$ denotes the observed set of links. This set is a subset $\mathcal{E}^o\subseteq \mathcal{E}^*$ of the complete true link set $\mathcal{E}^*\subseteq \mathcal{V}\times \mathcal{V}$, which includes not only the observed links but also all potential links that are currently unobserved or may appear in the future within $\mathcal{G}$. For any two nodes $u,v\in \mathcal{V}$, we denote the shortest-path between them as $\mathrm{SP}(u,v)$. The goal of link prediction tasks is to identify the set of unobserved true links $\mathcal{E}^u\subseteq \mathcal{E}^*\backslash \mathcal{E}^o$ in the given graph $\mathcal{G}$.

\paragraph{Tabular Foundation Model.} A tabular dataset is denoted as $\mathcal{T}=(\mathbf{X},\mathbf{y}^\prime)$, where $\mathbf{X}\in\mathbb{R}^{n\times d}$ contains $n$ rows and $d$ columns, and the label vector $\mathbf{y}^\prime\in\mathbb{R}^n$ offers supervision for a downstream task. A TFM is a pre-trained model $f(\cdot)$ designed to learn a universal inference mechanism, thereby enabling generalization to new and unseen tabular datasets without dataset-specific training. State-of-the-art TFMs, such as TabPFN~\cite{hollmanntabpfn,hollmann2025accurate} and LimiX~\cite{zhang2025limix}, accomplish this capability by first pre-training on a variety of synthetic datasets sampled from a carefully designed prior distribution, and then performing in-context Bayesian inference~\cite{liu2026tabularfoundationmodelsstrong}:
\begin{equation}
    p(\mathbf{y}^\prime_q|\mathbf{X}_q,\mathbf{X}_c,\mathbf{y}^\prime_c)\approx f(\mathbf{X}_q,\mathbf{X}_c,\mathbf{y}^\prime_c),
\end{equation}
where the pairs $(\mathbf{X}_q,\mathbf{y}^\prime_q)$ and $(\mathbf{X}_c,\mathbf{y}^\prime_c)$ refer to the test and in-context samples, respectively. 

An intuitive approach to adapt TFMs for link prediction is to treat each link as a table row and encode them into representations $\mathbf{Z}$. Observed positive links and randomly sampled negative links can serve as contexts $(\mathbf{Z}_c,\mathbf{y}_c)$. Given these contexts, TFMs perform in-context learning on the representation of target links $\mathbf{Z}_q$ to predict the existence labels:
\begin{equation}
    \mathbf{\hat{y}}_q=f(\mathbf{Z}_q,\mathbf{Z}_c,\mathbf{y}_c).
\end{equation}
However, this straight-forward approach faces two significant challenges. First, graphs for link prediction typically contain far more links than the context length capacity of TFMs can accommodate~\cite{ye2025closer,feuer2024tunetables}. An excessively large context not only fails to improve accuracy but also incurs substantial computational overhead. Randomly downsampling observed links can alleviate computational costs, but it also introduces sampling noise, causing the model to overfit to specific patterns within a single graph and neglect transferable patterns across different graphs in link prediction tasks. Second, it remains unclear how to obtain meaningful link representations $\mathbf{Z}$, since TFMs are pre-trained on tabular data and inherently lack the ability to capture complex graph topological information. %

\section{Method}\label{sec:method}

\begin{figure*}
    \centering
    \includegraphics[width=\textwidth]{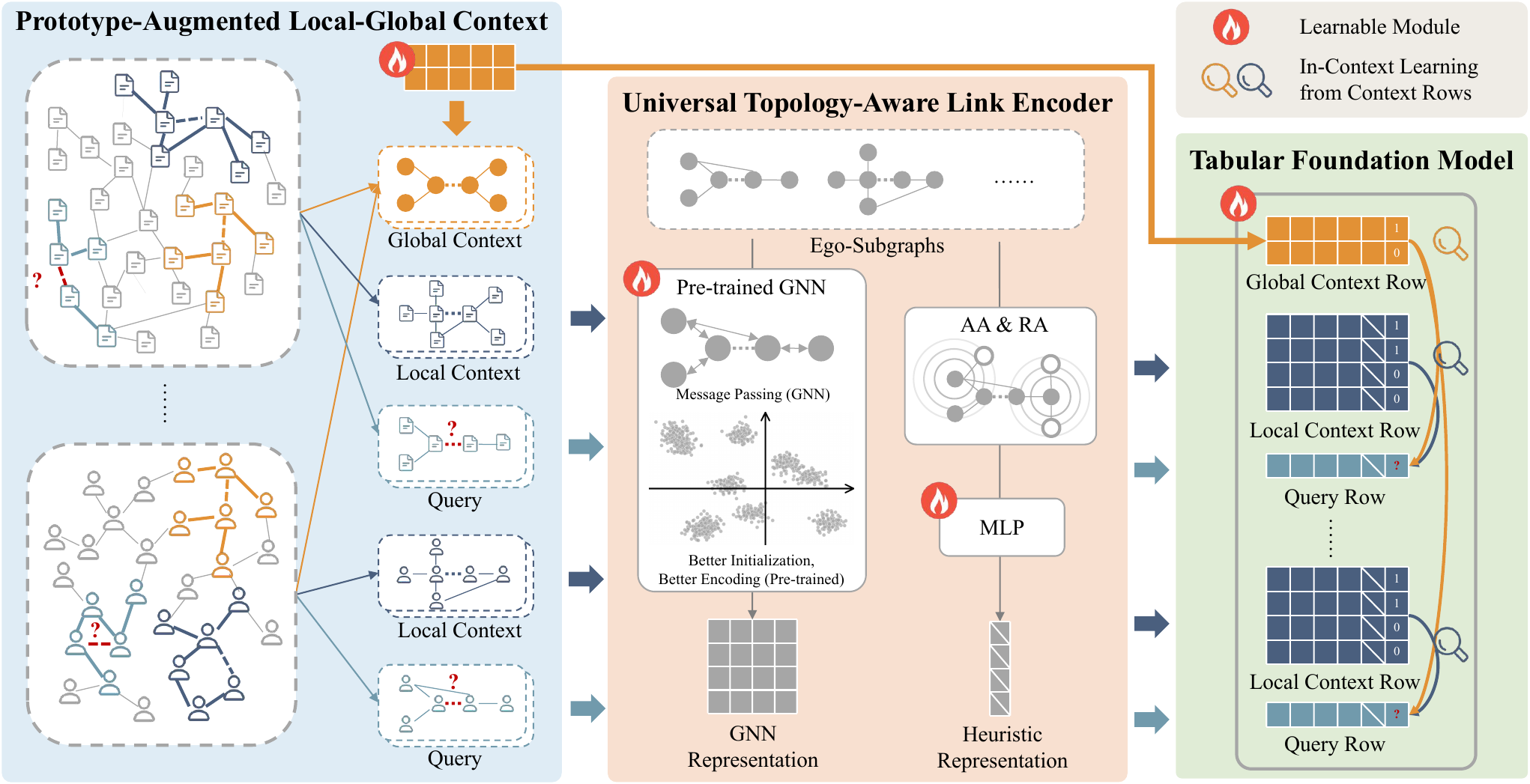}
    \caption{The framework of \name. Our proposed method is composed of three modules: (1) The \context~module constructs context through random subgraph downsampling and learnable prototypes, capturing both graph-specific and cross-graph transferable patterns. (2) The \encoder~module employs multiple link encoders to generate representations of context and target links. These representations provide rich and accurate link-centric topological information for the TFM. (3) The TFM performs in-context learning on the representations of context and target links to predict link existence labels.}
    \label{fig:framework}
\end{figure*}

In this section, we introduce \name. The overall framework is shown in Figure~\ref{fig:framework}. We first introduce the \context~module, which constructs context via random subgraph downsampling and learnable prototypes. Next, we present the \encoder, where multiple encoders generate representations for both context and target links. Finally, we describe how TFM performs in-context learning on the representations of context and targets to predict link existence labels.

\subsection{\Context}
The \context~module is designed to construct the context necessary for the TFMs' in-context learning setting. As discussed in Section~\ref{sec:form}, sampling context links solely from a single graph $\mathcal{G}$ faces the challenge of sampling noise, which can lead the model to overfit to task-specific patterns. In particular, the in-context learning performance of the model fluctuates with the sampled context links: performance tends to be better when the sampled context links are similar to the target links, and worse when they are dissimilar.

Therefore, a natural solution is to improve performance in cases where context links are dissimilar to the target links. Given the inductive biases commonly used in various link prediction tasks, such as the principle that node pairs sharing common neighbors are more likely to be linked~\cite{wangneural}, we hypothesize that there exist certain transferable patterns common to all link prediction tasks, even across different domains with distinct topological characteristics. If we can obtain additional context that represents such transferable patterns to complement the sampled context links, it would provide supplementary information, thereby improving the model’s in-context learning performance, especially when the sampled context links differ significantly from the target links. Accordingly, we refer to the sampled context links as the local context, and the context representing transferable patterns as the global context.

For the local context, we select $k$ connected node pairs $\mathcal{S}^+=\{s^+_1,s^+_2,\cdots,s^+_k\}\subseteq\mathcal{E}^o$ as positive samples. Similarly, we collect negative samples $\mathcal{S}^-=\{s^-_1,s^-_2,\cdots,s^-_k\}\subseteq\mathcal{V}\times\mathcal{V}\backslash\mathcal{E}^o$, comprising $k$ node pairs without link. The union $\mathcal{S}^+\cup\mathcal{S}^-$ approximates the overall properties of $\mathcal{G}$ while remaining computationally efficient. Once the positive and negative samples are obtained, we need to extract their structural representations. Following~\cite{zhang2018link,dong2025universal}, we start by extracting the ego-subgraph for each sample. For a node pair $s$, its ego-subgraph $\mathcal{G}^{\mathrm{ego}}_{s,r}$ is defined as the subgraph induced by all $r$-hop neighboring nodes of the nodes in $s$ on the original graph $\mathcal{G}$. For simplicity, we use 1-hop ego-subgraphs for all positive and negative samples $\mathcal{S}^+\cup\mathcal{S}^-$ to form the local context:
\begin{equation}
    \mathcal{C}_l=\{\mathcal{G}^{\mathrm{ego}}_{s,1}|s\in\mathcal{S}^+\cup\mathcal{S}^-\}.
\end{equation}
In addition, to ensure data consistency between the context and target links, we also construct the 1-hop ego-subgraph for the target link $q\in\mathcal{V}\times\mathcal{V}$, denoted as $\mathcal{G}^{\mathrm{ego}}_{q,1}$.

For the global context, inspired by prototype learning~\cite{lin2022prototypical}, we maintain a single set of learnable and shared prototypes $\mathbf{Z}_g\in\mathbb{R}^{p\times d}$ along with their associated labels $\mathbf{y}_g$ to capture the global task distribution across all link prediction tasks. Here, $p$ denotes the number of prototypes, $d$ is the dimension of the link representations, and the labels in $\mathbf{y}_g$ are evenly split between 0 and 1. By sharing the same prototypes across diverse tasks, we encourage them to learn transferable patterns. During both training and inference, these prototypes, referred to as the global context, are directly concatenated with the representations of the local context to form the input of the TFM described in Section~\ref{subsec:tfm}. 

The key distinction between our approach and other works that also employ learnable parameters as input to TFMs, often referred to as prompt tuning~\cite{ma2024context,feuer2024tunetables}, lies in their fundamental objectives. While prompt tuning aims to adapt the TFM to different downstream datasets, our prototypes are designed to capture the shared global context across all link prediction tasks, and they are not fine-tuned for each dataset. Consequently, their implementations also differ substantially. Prompt tuning typically requires a large number of learnable parameters (e.g., 1000 rows) to fit the distribution of a given downstream dataset. In contrast, the number of prototypes in our method is kept small (e.g., fewer than 100), as an excessive amount would shift the model's focus overly toward the global context and dilute dataset-specific characteristics. 

\subsection{\Encoder}

After obtaining the context using the \context~module, we address the fundamental task of transforming the graph-structured data into a two-dimensional, tabular format suitable for the TFM. This step is essential, as TFMs such as TabPFN~\cite{hollmanntabpfn,hollmann2025accurate} and LimiX~\cite{zhang2025limix} are inherently designed and pre-trained on tabular data, and therefore lack the capacity to capture complex graph topological information. To bridge this gap, we introduce a \encoder~designed to generate representations for both context and target links for the TFM to perform in-context learning.

Specifically, the representation of each context and target link is used to construct a row of a pseudo table, where each column corresponds to one dimension of the representation. The TFM then takes this pseudo table as input and performs in-context learning for the tabular task, predicting the label of the target row (i.e. the label of the target link) based on the labeled context rows and the unlabeled target row in Section~\ref{subsec:tfm}.
Crucially, the quality of the link representations used to construct the pseudo-table is essential: if the table itself is flawed or information-poor, the model cannot be expected to predict the target correctly. To obtain rich and accurate representations of both context and target links, we design our encoder to leverage multiple encoding approaches, including both learning-based methods such as GNNs and heuristic methods such as common neighbors, which have demonstrated strong potential in link prediction tasks~\cite{ma2024mixture}.

For the learning-based link encoder, we enhance the expressiveness of GNNs~\cite{hamilton2017inductive} in capturing graph structures by employing the labeling trick technique~\cite{zhang2018link,dong2025universal} to obtain link representations. Specifically, we first use DRNL+~\cite{dong2025universal}, a variant of Double Radius Node Labeling~\cite{zhang2018link} and Distance Encoding~\cite{li2020distance}, to generate structural features of each node in the ego-subgraph $\mathcal{G}^{\mathrm{ego}}_{e,1}$:
\begin{equation}
    \mathrm{DRNL+}(i,e)=\left\{
        \begin{aligned}
        &(0,\mathrm{SP}(i,u)), &\mathrm{if}\ \mathrm{SP}(i,v)=\infty, \\
        &(0,\mathrm{SP}(i,v)), &\mathrm{if}\ \mathrm{SP}(i,u)=\infty, \\
        &(\mathrm{DRNL(i,e)},0), &\mathrm{otherwise},
        \end{aligned}\right.
\end{equation}
where $e=(u,v)$ denotes a node pair from the target link or the context links, i.e., $e\in\{q\}\cup\mathcal{S}^+\cup\mathcal{S}^-$, and $\mathrm{DRNL(i,e)}$ is the integer label to node $i$ by Double Radius Node Labeling in $\mathcal{G}^{\mathrm{ego}}_{e,1}$. After the labeling trick encodes the relative structural information, we apply a pre-trained GNN $g(\cdot)$ from UniLP~\cite{dong2025universal} to update node representations. Following~\cite{zhang2018link,dong2025universal}, the ego-subgraph representation $h_e$ for each link $e$ is obtained by average pooling $\mathrm{Pool(\cdot)}$ over all node representations in $\mathcal{G}^{\mathrm{ego}}_{e,1}$:
\begin{equation}
    \mathbf{h}_e=\mathrm{Pool}(g(\mathbf{X}_{\mathrm{ego}},\mathbf{A}_{\mathrm{ego}})),
\end{equation}
where $\mathbf{X}_{\mathrm{ego}}$ is the node features formed by $\mathrm{DRNL+}(i,e)$ for each node $i$ in $\mathcal{G}^{\mathrm{ego}}_{e,1}$, and $\mathbf{A}_{\mathrm{ego}}$ is the adjacency matrix of $\mathcal{G}^{\mathrm{ego}}_{e,1}$. We employ a pre-trained GNN instead of a randomly initialized one because the latter tends to overfit the specific in-context learning patterns of the TFM during early training stages, which results in performance degradation due to reduced model expressiveness. Using a pre-trained GNN effectively alleviates this issue, allowing the TFM to conduct in-context learning based on richer representations.

For the heuristic methods, we consider Adamic-Ada (AA)~\cite{adamic2003friends} and Resource Allocation (RA)~\cite{zhou2009predicting} to provide additional structural information. This is motivated by the fact that the 1-hop ego-subgraph captures only local connectivity, and thus lacks awareness of structural patterns beyond the immediate neighborhood. Specifically, we apply an MLP $w(\cdot)$ to generate the link representation from the heuristic methods for the node pair $e$:
\begin{equation}
    \mathbf{h}^\prime_e=w([\mathrm{AA(e),RA(e)}]).
\end{equation}
After obtaining link representations from both the GNN and heuristic methods, we concatenate these representations to form the final link representation $\mathbf{z}_e = [\mathbf{h}_e, \mathbf{h}^\prime_e] \in \mathbb{R}^d$ for link $e$, where $d$ denotes the dimensionality. %
The local context representations $\mathbf{Z}_l$ are then obtained by concatenating the representation $\mathbf{z}_s$ of each local context $s\in\mathcal{S}^+\cup\mathcal{S}^-$. Similarly, we derive the representation $\mathbf{z}_q$ of the target link $q$. These representations are used as inputs for the TFM.

\subsection{Tabular Foundation Model}\label{subsec:tfm}
In this section, we introduce how TFM performs in-context learning on the representations of context and target links to predict link existence labels. First, we combine the local and global context representations and ground-truth labels to obtain the final context representations and context labels:
\begin{equation}
    \mathbf{Z}_c=[\mathbf{Z}_l,\mathbf{Z}_g],\mathbf{y}_c=[\mathbf{y}_l,\mathbf{y}_g],
\end{equation}
where $\mathbf{y}_l$ and $\mathbf{y}_g$ are the ground-truth labels of the local and global context, respectively. Next, we use the TFM $f(\cdot)$, such as TabPFN~\cite{hollmann2025accurate} and LimiX~\cite{zhang2025limix}, to perform tabular-based in-context inference:
\begin{equation}
    \hat{y}_q=f(\mathbf{z}_q,\mathbf{Z}_c,\mathbf{y}_c).
\end{equation}

To adapt \name~to diverse link prediction tasks, we perform full fine-tuning of the entire model without employing other complex optimization strategies, as full fine-tuning has been shown to achieve the fastest convergence among fine-tuning methods for such TFMs~\cite{rubachev2025finetuningtabularfoundationmodels}. Given a set of graphs $\{\mathcal{G}_1,\mathcal{G}_2,\cdots,\mathcal{G}_n\}$ from different tasks, where each graph $\mathcal{G}$ belongs to this set, the optimization objective is defined as follows:
\begin{equation}
    \min_\Theta\mathbb{E}_{\mathcal{G}\in\{\mathcal{G}_1,\mathcal{G}_2,\cdots,\mathcal{G}_n\},q\in\mathcal{V}\times\mathcal{V}}[\ell(\hat{y}_q,y_q)],
\end{equation}
where $\Theta$ are all learnable parameters, including those of the prototypes $\mathbf{Z}_g$, the GNN $g(\cdot)$, the MLP $w(\cdot)$, and the TFM $f(\cdot)$. $\ell(\cdot)$ denotes the loss function on a single instance, and $y_q$ is the ground-truth existence label of link $q$.

\subsection{Discussions}
The time complexity of our \name~is $O(k\overline{m}_ed_1+k\overline{m}_vd^2_1+(2k+p)dd_2(2k+p+d+d_2))$, where $k$ denotes the number of positive or negative context samples, $p$ is the number of learnable prototypes, $\overline{m}_e$ is the average number of links in an ego-subgraph, $\overline{m}_v$ is the average number of nodes in an ego-subgraph, $d$ is the dimensionality of the link representation produced by the GNN and heuristic methods, $d_1$ is the feature dimension of the GNN, and $d_2$ is the encoding dimension of the TFM. Specifically, extracting the $2k+1$ 1‑hop ego-subgraphs (including context and target links) requires $O(k\overline{m}_e)$ operations, while the GNN encoding of these subgraphs costs $O(k\overline{m}_ed_1+k\overline{m}_vd^2_1)$. Heuristic feature computation, which can be pre‑computed in batch for the entire graph, incurs negligible overhead. Within the TFM, the feature‑level attention mechanism contributes $O((2k+p)d^2d_2)$, the sample‑level attention mechanism contributes $O((2k+p)^2dd_2)$, and the feed‑forward network adds $O((2k+p)dd^2_2)$. Summing these components yields the final complexity expression. Notably, the computational cost outside the TFM remains consistent with typical GNN-based ego-subgraph approaches such as SEAL~\cite{zhang2018link}. Although incorporating the TFM introduces additional overhead for in‑context learning, it effectively eliminates the need for computationally expensive fine‑tuning on downstream tasks.

\section{Experiments}\label{sec:exp}

\begin{table*}
  \caption{Performance comparison with competitive link prediction methods on test datasets evaluated by Hits@50. Numbers after $\pm$ indicate variances. The best results are in \best{bold} and the second-best results are \secbest{underlined}.}
  \begin{tabular}{lccccccccc}
    \toprule
    &Biology&Transport&Web&\multicolumn{2}{c}{Collaboration}&Citation&Social&\\
    &C.ele&USAir&PB&NS&CS&Cora&Facebook&Average&Ave. Rank\\
    \midrule
    \multicolumn{9}{l}{\textit{Heuristics}}\\
    CN&46.88\scalebox{0.75}{±12.28}&82.75\scalebox{0.75}{±1.54}&41.15\scalebox{0.75}{±3.77}&74.03\scalebox{0.75}{±1.59}&56.84\scalebox{0.75}{±15.56}&33.85\scalebox{0.75}{±0.93}&58.70\scalebox{0.75}{±0.35}&56.31&12.29\\
    AA&61.07\scalebox{0.75}{±5.16}&86.96\scalebox{0.75}{±2.24}&44.12\scalebox{0.75}{±3.36}&74.03\scalebox{0.75}{±1.59}&68.22\scalebox{0.75}{±1.08}&33.85\scalebox{0.75}{±0.93}&67.80\scalebox{0.75}{±2.12}&62.29&7.71\\
    RA&62.80\scalebox{0.75}{±4.84}&87.27\scalebox{0.75}{±1.89}&43.72\scalebox{0.75}{±2.86}&74.03\scalebox{0.75}{±1.59}&68.21\scalebox{0.75}{±1.08}&33.85\scalebox{0.75}{±0.93}&\secbest{68.84}\scalebox{0.75}{±2.03}&62.67&7.29\\
    PA&43.85\scalebox{0.75}{±4.12}&77.69\scalebox{0.75}{±2.29}&28.93\scalebox{0.75}{±1.91}&35.35\scalebox{0.75}{±3.01}&6.49\scalebox{0.75}{±0.61}&22.09\scalebox{0.75}{±1.52}&12.95\scalebox{0.75}{±0.63}&32.48&15.71\\
    SP&0.00\scalebox{0.75}{±0.00}&0.00\scalebox{0.75}{±0.00}&0.00\scalebox{0.75}{±0.00}&80.00\scalebox{0.75}{±1.11}&41.34\scalebox{0.75}{±35.58}&52.97\scalebox{0.75}{±1.53}&0.00\scalebox{0.75}{±0.00}&24.90&15.00\\
    Katz&58.86\scalebox{0.75}{±6.48}&84.64\scalebox{0.75}{±1.86}&44.36\scalebox{0.75}{±3.65}&78.96\scalebox{0.75}{±1.35}&66.32\scalebox{0.75}{±5.59}&52.97\scalebox{0.75}{±1.53}&60.79\scalebox{0.75}{±0.60}&63.84&8.86\\
    \midrule
    \multicolumn{9}{l}{\textit{Pre-train Only GNN}}\\
    SEAL&61.28\scalebox{0.75}{±3.76}&86.00\scalebox{0.75}{±1.56}&45.44\scalebox{0.75}{±2.68}&84.07\scalebox{0.75}{±1.96}&62.82\scalebox{0.75}{±1.62}&56.21\scalebox{0.75}{±2.24}&54.57\scalebox{0.75}{±1.48}&64.34&7.57\\
    GAE&44.71\scalebox{0.75}{±3.39}&76.12\scalebox{0.75}{±2.27}&27.56\scalebox{0.75}{±2.34}&15.20\scalebox{0.75}{±2.14}&5.08\scalebox{0.75}{±0.48}&24.22\scalebox{0.75}{±1.53}&6.65\scalebox{0.75}{±0.57}&28.51&16.14\\
    ELPH&59.23\scalebox{0.75}{±4.50}&84.42\scalebox{0.75}{±2.22}&43.69\scalebox{0.75}{±2.90}&84.27\scalebox{0.75}{±1.43}&\secbest{70.69}\scalebox{0.75}{±3.63}&56.91\scalebox{0.75}{±1.43}&61.80\scalebox{0.75}{±2.46}&65.86&7.43\\
    NCNC&48.07\scalebox{0.75}{±4.79}&75.44\scalebox{0.75}{±4.22}&25.66\scalebox{0.75}{±1.42}&80.07\scalebox{0.75}{±1.43}&34.27\scalebox{0.75}{±2.28}&52.51\scalebox{0.75}{±2.54}&19.28\scalebox{0.75}{±1.58}&47.90&13.57\\
    MPLP&56.74\scalebox{0.75}{±5.31}&82.94\scalebox{0.75}{±2.30}&47.78\scalebox{0.75}{±2.55}&80.33\scalebox{0.75}{±1.54}&24.26\scalebox{0.75}{±1.28}&46.71\scalebox{0.75}{±2.25}&48.06\scalebox{0.75}{±2.09}&55.26&10.71\\
    \midrule
    \multicolumn{9}{l}{\textit{Pre-train \& Fine-tune GNN}}\\
    SEAL&64.45\scalebox{0.75}{±4.14}&\secbest{88.49}\scalebox{0.75}{±2.16}&47.78\scalebox{0.75}{±3.32}&84.84\scalebox{0.75}{±2.32}&61.54\scalebox{0.75}{±3.09}&\secbest{62.19}\scalebox{0.75}{±3.27}&58.70\scalebox{0.75}{±2.78}&66.86&4.71\\
    GAE&44.71\scalebox{0.75}{±4.07}&74.47\scalebox{0.75}{±2.96}&25.92\scalebox{0.75}{±2.64}&18.34\scalebox{0.75}{±2.27}&4.95\scalebox{0.75}{±0.44}&25.31\scalebox{0.75}{±1.48}&6.11\scalebox{0.75}{±0.39}&28.54&16.57\\
    ELPH&60.51\scalebox{0.75}{±5.72}&84.52\scalebox{0.75}{±2.08}&43.58\scalebox{0.75}{±3.48}&86.08\scalebox{0.75}{±0.69}&\best{71.10}\scalebox{0.75}{±3.48}&57.18\scalebox{0.75}{±1.89}&63.31\scalebox{0.75}{±3.63}&66.61&6.57\\
    NCNC&64.45\scalebox{0.75}{±5.10}&85.85\scalebox{0.75}{±2.46}&47.75\scalebox{0.75}{±6.90}&88.25\scalebox{0.75}{±2.25}&58.75\scalebox{0.75}{±5.74}&60.00\scalebox{0.75}{±2.50}&59.32\scalebox{0.75}{±6.93}&66.34&5.57\\
    MPLP&62.56\scalebox{0.75}{±4.79}&85.08\scalebox{0.75}{±1.54}&\secbest{48.01}\scalebox{0.75}{±2.94}&80.16\scalebox{0.75}{±1.18}&50.35\scalebox{0.75}{±1.01}&56.02\scalebox{0.75}{±2.09}&56.72\scalebox{0.75}{±1.20}&62.70&8.00\\
    \midrule
    \multicolumn{9}{l}{\textit{GFM}}\\
    UniLP&\secbest{65.20}\scalebox{0.75}{±4.40}&85.98\scalebox{0.75}{±2.00}&\best{48.14}\scalebox{0.75}{±2.99}&\secbest{89.09}\scalebox{0.75}{±2.05}&64.59\scalebox{0.75}{±2.65}&57.50\scalebox{0.75}{±2.40}&65.49\scalebox{0.75}{±2.05}&\secbest{68.00}&\secbest{3.71}\\
    \name&\best{68.62}\scalebox{0.75}{±5.97}&\best{91.67}\scalebox{0.75}{±1.37}&47.75\scalebox{0.75}{±2.76}&\best{90.46}\scalebox{0.75}{±2.43}&70.42\scalebox{0.75}{±1.91}&\best{62.63}\scalebox{0.75}{±1.77}&\best{69.06}\scalebox{0.75}{±1.12}&\best{71.52}&\best{1.86}\\
    \bottomrule
  \end{tabular}
  \label{tab:performance}
\end{table*}

\begin{figure*}
\centering
	\begin{subfigure}{0.24\linewidth}
		\centering
		\includegraphics[width=\linewidth]{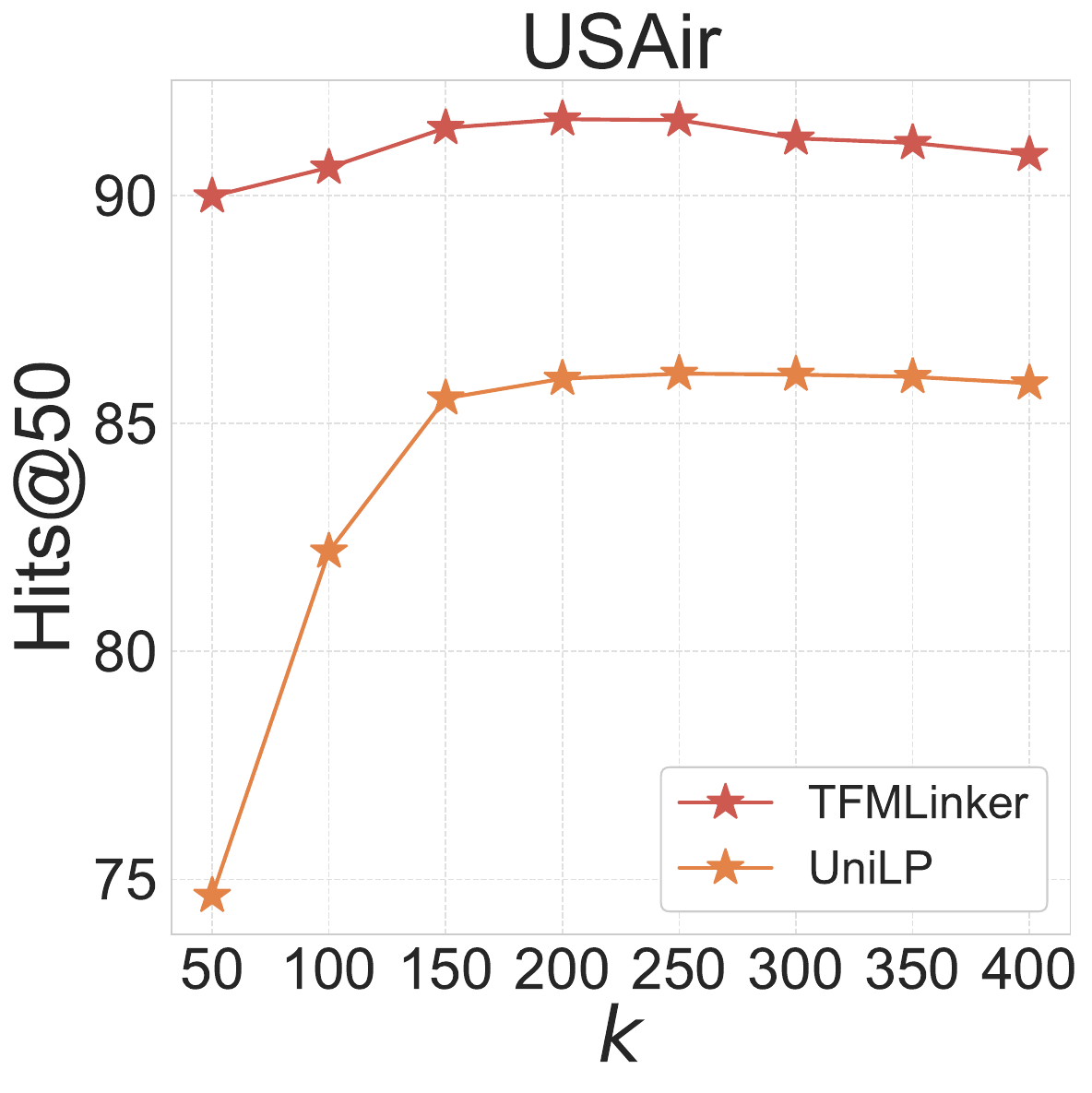}
	\end{subfigure}
	\begin{subfigure}{0.24\linewidth}
		\centering
		\includegraphics[width=\linewidth]{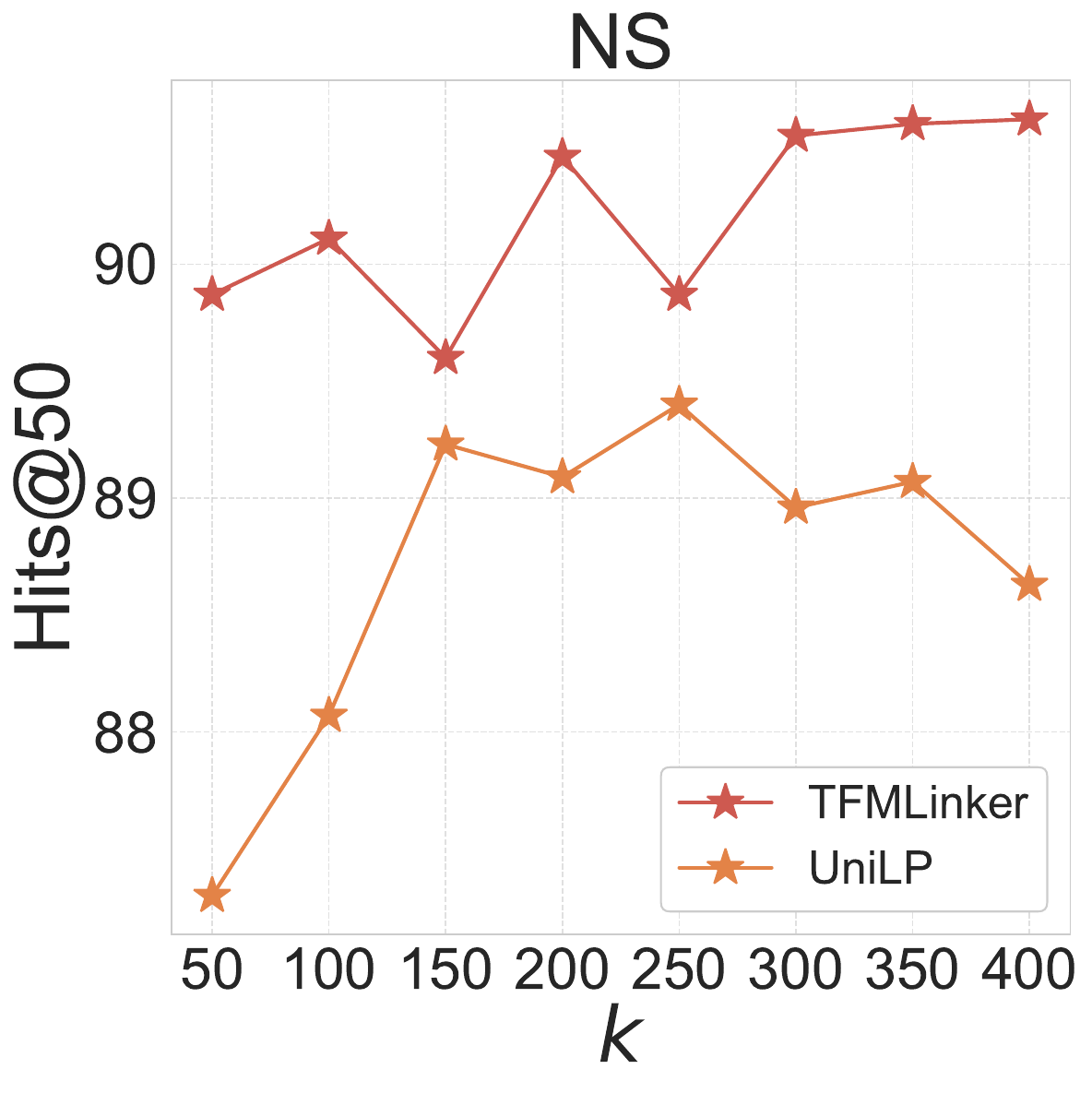}
	\end{subfigure}
	\begin{subfigure}{0.24\linewidth}
		\centering
		\includegraphics[width=\linewidth]{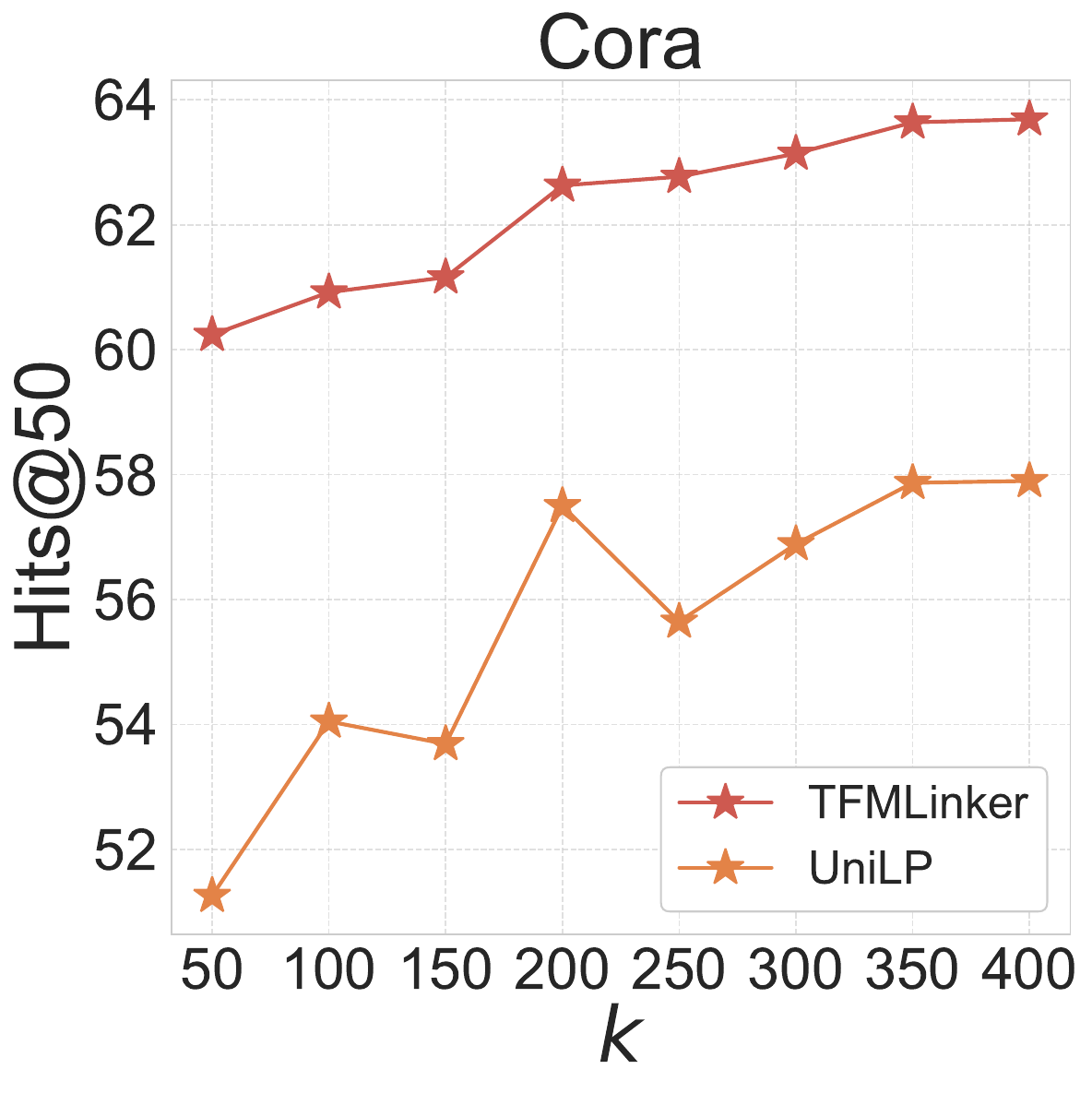}
	\end{subfigure}
	\begin{subfigure}{0.24\linewidth}
		\centering
		\includegraphics[width=\linewidth]{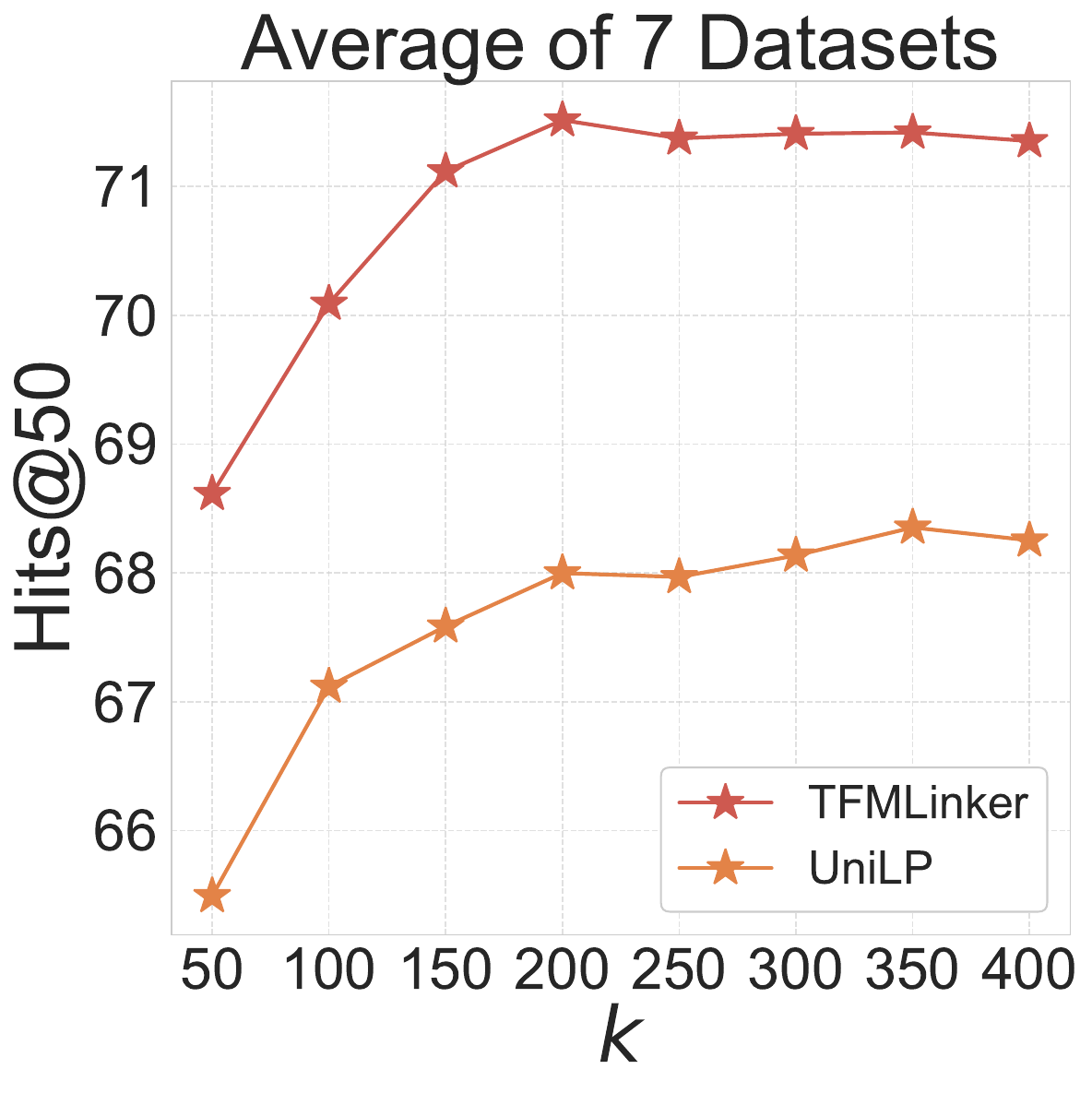}
	\end{subfigure}
    \vspace{-0.3cm}
    \caption{The impact of the number of in-context links $k$. }%
    \label{fig:icl_num}
\end{figure*}

In this section, we conduct extensive experiments to evaluate \name, focusing on three questions:
\begin{itemize}[leftmargin=0.5cm]
    \item \textbf{RQ1}: Can \name~effectively improve universal link prediction abilities across datasets and domains without fine-tuning?
    \item \textbf{RQ2}: How effective is \name~under different quantities of in-context links and with different TFMs?
    \item \textbf{RQ3}: How do the key components and hyper-parameters influence the results?
\end{itemize}

\subsection{Experimental Setup}

\textbf{Datasets.} The training of \name~is based on a collection of graph datasets spanning diverse domains. Following~\cite{maorevisiting,dong2025universal}, we utilize graph data from fields such as biology~\cite{von2002comparative,watts1998collective,zhang2018beyond}, transportation~\cite{watts1998collective}, the web~\cite{ackland2005mapping,adamic2005political,spring2002measuring}, academic collaboration~\cite{newman2006finding,shchur2018pitfalls}, citation networks~\cite{yang2016revisiting}, and social networks~\cite{rozemberczki2021multi}. This diverse selection ensures that our training and evaluation of link prediction models cover a wide spectrum of connectivity patterns. For training, we use datasets such as Ecoli, Yeast, Power, PolBlogs, Router, Physics, Pubmed, Citeseer, Twitch, and Github. For evaluation, the test set comprises Celegans, USAir, PB, NS, CS, Cora, and Facebook. Details of the curated graph datasets are provided in Appendix~\ref{subsec:dataset-detail}.

\noindent\textbf{Baselines.} We compare \name~with state-of-the-art link prediction methods from three groups. The first group consists of heuristic methods, including Common Neighbor (CN)~\cite{liben2003link}, Adamic-Adar (AA)~\cite{adamic2003friends}, Resource Allocation (RA)~\cite{zhou2009predicting}, Preferential Attachment (PA)~\cite{barabasi1999emergence}, Shortest-Path (SP), and Katz index (Katz)~\cite{katz1953new}. The second group is GNN-based parametric link predictors that include GAE~\cite{kipf2016variational}, SEAL~\cite{zhang2018link}, ELPH~\cite{chamberlaingraph}, NCNC~\cite{wangneural}, and MPLP~\cite{dong2024pure}. The third group includes a GFM for cross-domain link prediction UniLP~\cite{dong2025universal}. Following the setup in~\cite{dong2025universal}, we provide GAE and NCNC with a 32-dimensional all-one vector as initial node features, as they require attributed input. All other methods operate directly on the non-attributed graph structure.

\noindent\textbf{Evaluation.} Following the evaluation protocol of~\cite{dong2025universal}, we assess the in-context learning performance of UniLP and \name~on unseen datasets. The graph data is partitioned into non-overlapping training and test sets. A single model is trained on the combined training datasets, during which 40 positive and 40 negative links are dynamically sampled from each corresponding training dataset to serve as in-context links for every target link. For evaluation, each test dataset is further divided into training, validation, and test subsets in a 70\%:10\%:20\% ratio. The training subset corresponds to observed links, whereas the validation and test subsets represent unobserved links. During inference, we sample 200 positive and 200 negative links from each test dataset to form the in-context set. Other baselines follow a comparable protocol. For GNNs, we adopt two experimental settings. In the first setting, termed Pre-train Only, models are trained on the combined pretraining datasets and then directly evaluated on each test dataset. In the second setting, referred to as Pre-train \& Fine-tune, models are first pretrained and then additionally fine-tuned on each test dataset using 200 sampled positive and 200 sampled negative links for training.

\begin{table*}
  \caption{The results of \name~with different backbones. The best results are in \best{bold} and the runner-ups are \secbest{underlined}.}
  \begin{tabular}{lcccccccc}
    \toprule
    &Biology&Transport&Web&\multicolumn{2}{c}{Collaboration}&Citation&Social&\\
    &C.ele&USAir&PB&NS&CS&Cora&Facebook&Average\\
    \midrule
    SEAL&64.45\scalebox{0.75}{±4.14}&88.49\scalebox{0.75}{±2.16}&47.78\scalebox{0.75}{±3.32}&84.84\scalebox{0.75}{±2.32}&61.54\scalebox{0.75}{±3.09}&62.19\scalebox{0.75}{±3.27}&58.70\scalebox{0.75}{±2.78}&66.86\\
    UniLP&65.20\scalebox{0.75}{±4.40}&85.98\scalebox{0.75}{±2.00}&\best{48.14}\scalebox{0.75}{±2.99}&\secbest{89.09}\scalebox{0.75}{±2.05}&64.59\scalebox{0.75}{±2.65}&57.50\scalebox{0.75}{±2.40}&65.49\scalebox{0.75}{±2.05}&68.00\\
    \midrule
    \name~+ TabPFNv2&\secbest{67.86}\scalebox{0.75}{±6.59}&\best{92.09}\scalebox{0.75}{±1.18}&\secbest{47.91}\scalebox{0.75}{±2.23}&86.19\scalebox{0.75}{±1.43}&\best{70.77}\scalebox{0.75}{±1.56}&61.39\scalebox{0.75}{±1.85}&\best{69.92}\scalebox{0.75}{±1.57}&\secbest{70.88}\\
    \name~+ Limix-2M&65.66\scalebox{0.75}{±5.31}&91.15\scalebox{0.75}{±1.46}&47.03\scalebox{0.75}{±3.03}&87.77\scalebox{0.75}{±1.82}&70.13\scalebox{0.75}{±1.87}&\best{63.43}\scalebox{0.75}{±1.60}&\secbest{69.80}\scalebox{0.75}{±1.17}&70.71\\
    \name~+ Limix-16M&\best{68.62}\scalebox{0.75}{±5.97}&\secbest{91.67}\scalebox{0.75}{±1.37}&47.75\scalebox{0.75}{±2.76}&\best{90.46}\scalebox{0.75}{±2.43}&\secbest{70.42}\scalebox{0.75}{±1.91}&\secbest{62.63}\scalebox{0.75}{±1.77}&69.06\scalebox{0.75}{±1.12}&\best{71.52}\\
    \bottomrule
  \end{tabular}
  \label{tab:tfm-selection}
\end{table*}

\noindent\textbf{Implementation Details.} We employ a 3-layer GraphSAGE~\cite{hamilton2017inductive} with 128-dimensional hidden representations as the GNN and employ the LimiX-16M~\cite{zhang2025limix} as the default TFM. The heuristic-based link representations have a dimensionality of 16. The number of learnable prototypes is set to 80. No dropout is applied. The model is trained for 5 epochs using a learning rate of 1e-5. For each target link, we sample 40 positive and 40 negative links from the corresponding training dataset as in-context links. This design introduces variability: different target links from the same dataset, or the same target link across training batches, may be associated with different sets of in-context links. All experimental results are reported as mean ± standard deviation over 10 independent runs. Additional implementation details are provided in Appendix~\ref{subsec:implementation-detail}.

\subsection{Performance Comparisons}

To answer \textbf{RQ1}, we report the results of \name~compared to baseline methods on various unseen graph datasets in Table~\ref{tab:performance}. %
The results show that \name~consistently outperforms traditional heuristic methods, standard GNN-based link prediction models, and the state-of-the-art in-context learning GFM UniLP, achieving improvements on 5 out of 7 benchmark datasets. It also demonstrates clear advantages in terms of overall ranking and average metrics over all baselines. For example, on the C.ele dataset, \name~achieves a 5.25\% improvement in Hits@50 over the second-best method. Moreover, compared to the best-performing baseline, our method advances the overall ranking by 1.85 positions and raises the average metric by 5.18\%.

Furthermore, \name~demonstrates strong generalization across a wide range of unseen graph datasets, achieving performance comparable or even superior to fine‑tuned GNN‑based link prediction models, despite not being explicitly trained on the test dataset. This capability stems from its in‑context learning mechanism based on TFMs, which enables the model to adapt on-the-fly to specific graph datasets without additional training. By leveraging in‑context links during inference, \name~dynamically adjusts its understanding of connectivity patterns, resulting in generalizable and flexible performance.

\subsection{In-Depth Analyses}
\subsubsection{Adaptability to Context Length}

To answer \textbf{RQ2}, we conduct experiments with varying in-context link numbers $k$, sampled from each test graph in the range of 50 to 400. The results on the USAir, NS, Cora datasets, and the average performance across all seven test datasets are presented in Figure~\ref{fig:icl_num}. These sampled links serve as contextual input for the model, with UniLP employed as the base model. The findings show that \name’s performance consistently improves as the number of in-context links increases, indicating that our method can more effectively capture the properties of the target graph when provided with richer contextual information. Moreover, regardless of $k$, \name~consistently achieves average performance superior to the best baseline, demonstrating the generalization of its in-context learning ability across different numbers of contextual samples.

\subsubsection{Impact of TFM Backbone}

We further evaluate \name~with various TFMs, including TabPFNv2, Limix-2M and Limix-16M, as presented in Table~\ref{tab:tfm-selection}. For comparison, we utilize SEAL (under the Pre-train \& Fine-tune setting) and UniLP as the base models, given their status as two of the best-performing baselines. The results demonstrate that \name~consistently surpasses these strong baselines on 6 out of 7 benchmark datasets, showing superior overall performance across different TFMs. This further confirms the effectiveness of \name, irrespective of the specific TFM employed. Furthermore, as the TFM size increases from Limix-2M to TabPFNv2 and then to Limix-16M, the average performance metrics of \name~show consistent improvement. This demonstrates that \name~can continuously benefit from the enhanced capabilities of larger TFMs.

\subsubsection{Ablation Studies}\label{subsec:ablation}

\begin{table*}
  \caption{The results of ablation studies on \name. \textit{w/o global context} denotes the variant with only local context; \textit{w/o GNN pre-train} uses randomly initialized GNN weights; \textit{w/o GNN module} removes the GNN module entirely; \textit{w/o heuristics} excludes all heuristic features; \textit{w/o TFM pre-train} replaces the pre-trained TFM with a randomly initialized one. The best results are in \best{bold} and the second-best results are \secbest{underlined}.}
  \begin{tabular}{lcccccccc}
    \toprule
    &Biology&Transport&Web&\multicolumn{2}{c}{Collaboration}&Citation&Social&\\
    &C.ele&USAir&PB&NS&CS&Cora&Facebook&Average\\
    \midrule
    \name&\secbest{68.62}\scalebox{0.75}{±5.97}&\best{91.67}\scalebox{0.75}{±1.37}&\secbest{47.75}\scalebox{0.75}{±2.76}&\best{90.46}\scalebox{0.75}{±2.43}&\best{70.42}\scalebox{0.75}{±1.91}&\secbest{62.63}\scalebox{0.75}{±1.77}&69.06\scalebox{0.75}{±1.12}&\best{71.52}\\
    \textit{w/o global context}&66.97\scalebox{0.75}{±5.34}&90.59\scalebox{0.75}{±1.83}&45.23\scalebox{0.75}{±3.00}&\secbest{88.74}\scalebox{0.75}{±1.76}&62.10\scalebox{0.75}{±7.75}&61.68\scalebox{0.75}{±1.58}&\secbest{69.27}\scalebox{0.75}{±1.51}&69.23\\
    \textit{w/o GNN pre-train}&\best{68.95}\scalebox{0.75}{±6.66}&\secbest{91.20}\scalebox{0.75}{±1.81}&47.56\scalebox{0.75}{±2.17}&81.95\scalebox{0.75}{±2.55}&68.08\scalebox{0.75}{±2.46}&60.93\scalebox{0.75}{±1.49}&\best{70.49}\scalebox{0.75}{±1.64}&69.88\\
    \textit{w/o GNN module}&62.17\scalebox{0.75}{±6.29}&88.52\scalebox{0.75}{±1.33}&44.84\scalebox{0.75}{±2.19}&77.79\scalebox{0.75}{±3.17}&68.81\scalebox{0.75}{±0.96}&38.12\scalebox{0.75}{±3.42}&68.89\scalebox{0.75}{±0.73}&64.16\\
    \textit{w/o heuristics}&67.65\scalebox{0.75}{±5.76}&90.24\scalebox{0.75}{±1.50}&\best{48.48}\scalebox{0.75}{±2.83}&88.32\scalebox{0.75}{±1.73}&\secbest{69.27}\scalebox{0.75}{±2.34}&\best{63.18}\scalebox{0.75}{±1.53}&67.52\scalebox{0.75}{±1.82}&\secbest{70.67}\\
    \textit{w/o TFM pre-train}&66.92\scalebox{0.75}{±6.64}&87.39\scalebox{0.75}{±2.12}&37.84\scalebox{0.75}{±5.08}&85.82\scalebox{0.75}{±1.59}&59.36\scalebox{0.75}{±1.35}&50.67\scalebox{0.75}{±2.74}&62.25\scalebox{0.75}{±1.27}&64.32\\
    \bottomrule
  \end{tabular}
  \label{tab:ablation}
  \vspace{0.3cm}
\end{table*}

\begin{figure*}
\centering
	\begin{subfigure}{0.24\linewidth}
		\centering
		\includegraphics[width=\linewidth]{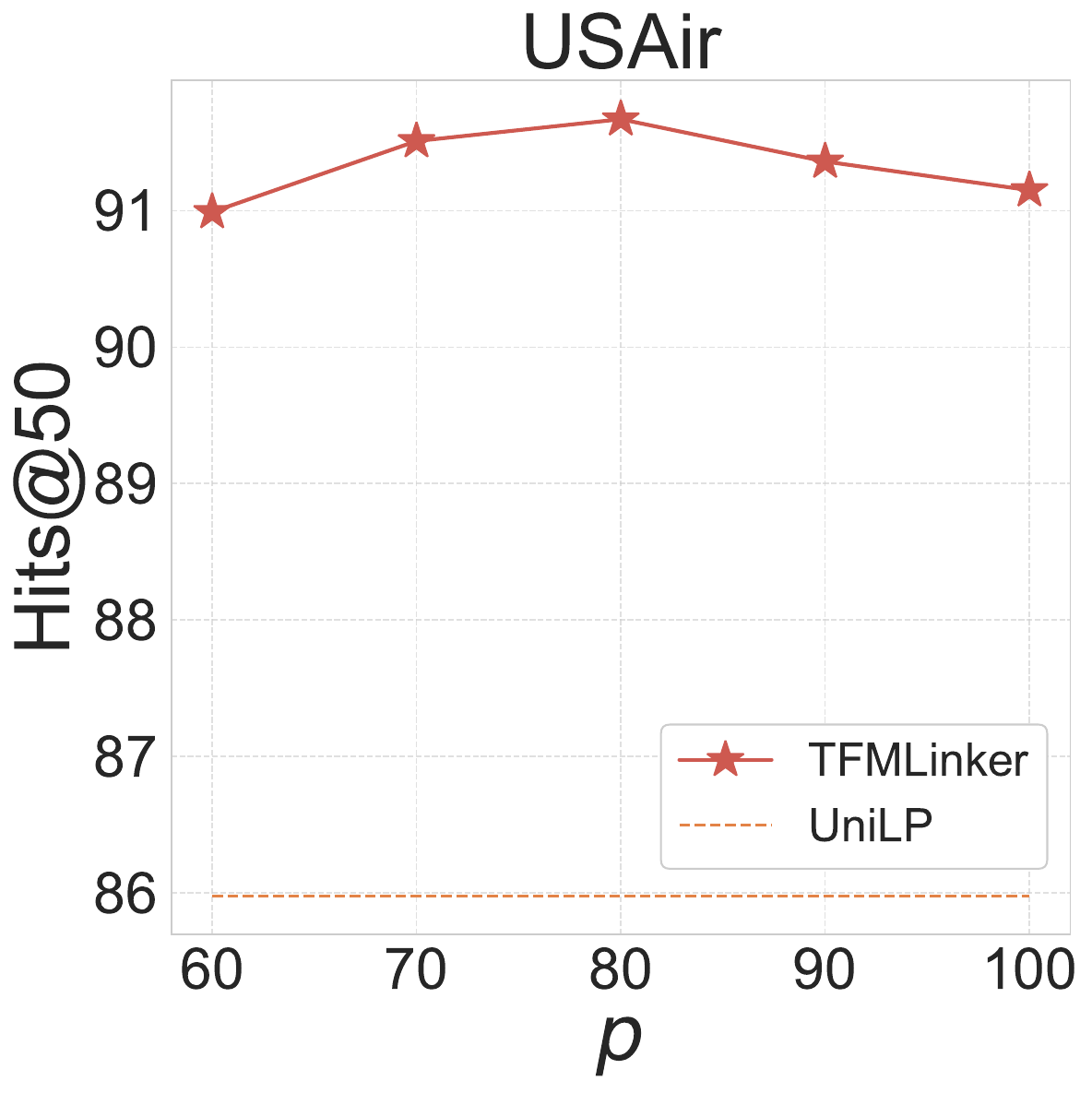}
	\end{subfigure}
	\begin{subfigure}{0.24\linewidth}
		\centering
		\includegraphics[width=\linewidth]{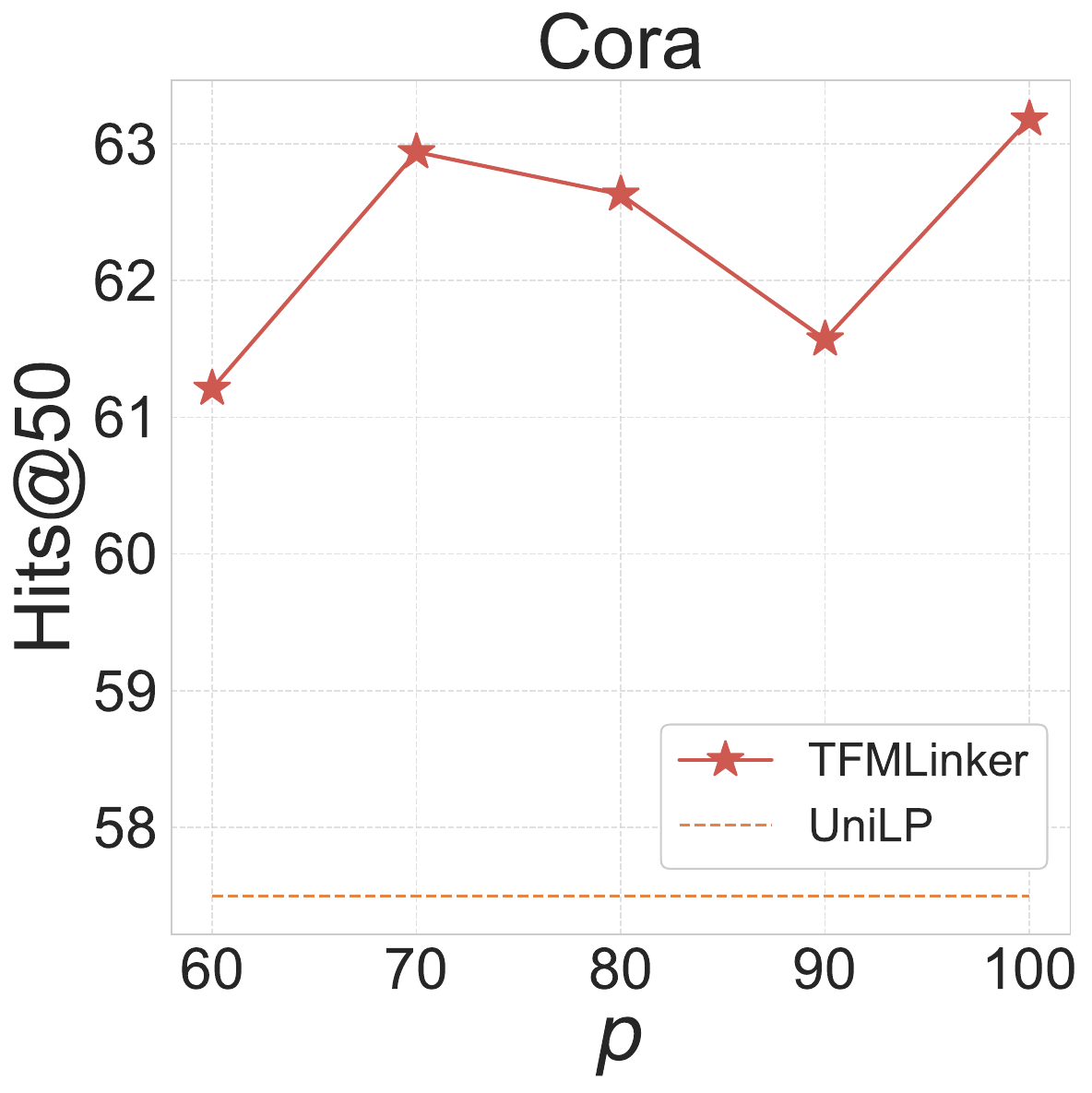}
	\end{subfigure}
	\begin{subfigure}{0.24\linewidth}
		\centering
		\includegraphics[width=\linewidth]{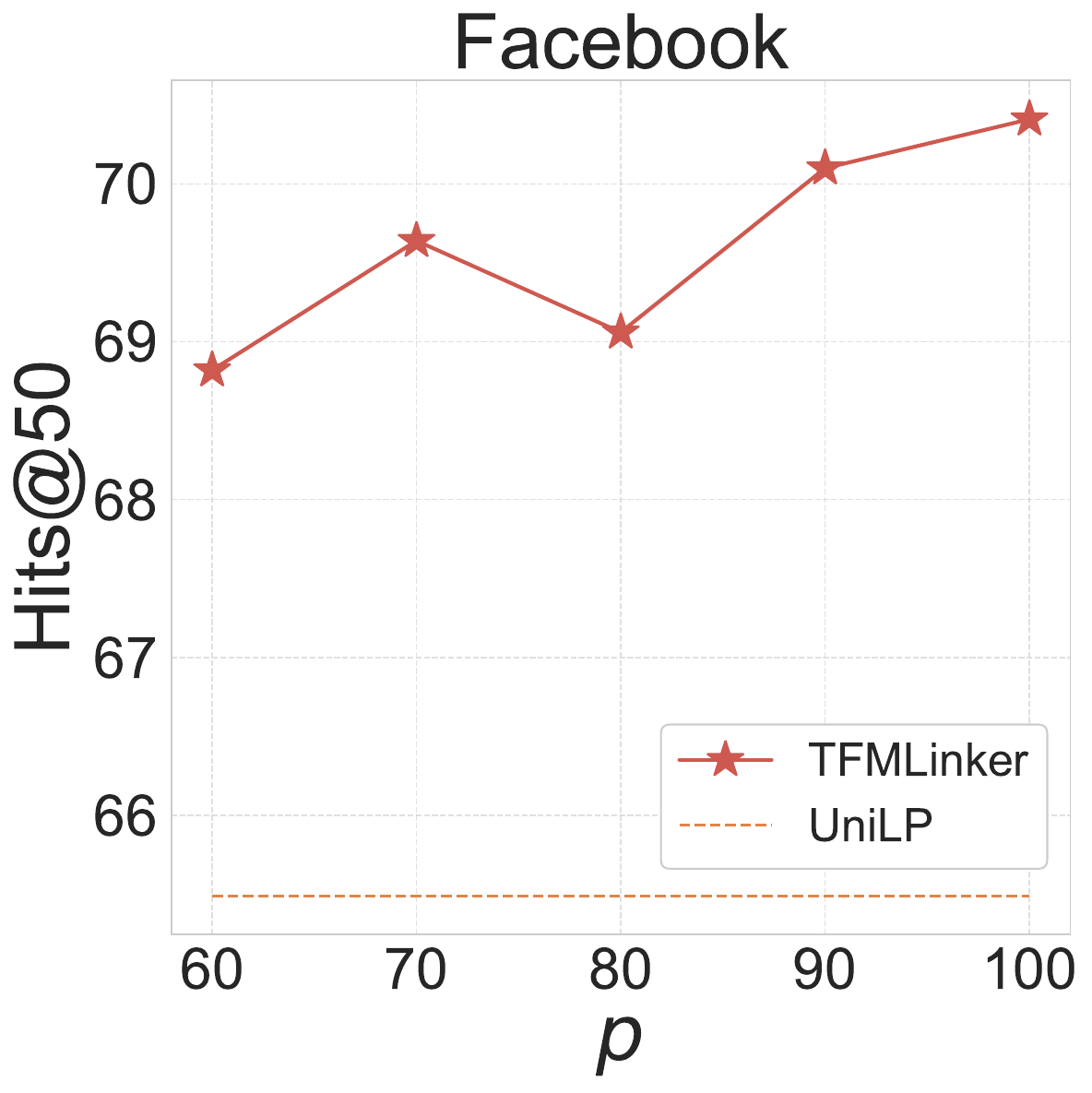}
	\end{subfigure}
	\begin{subfigure}{0.24\linewidth}
		\centering
		\includegraphics[width=\linewidth]{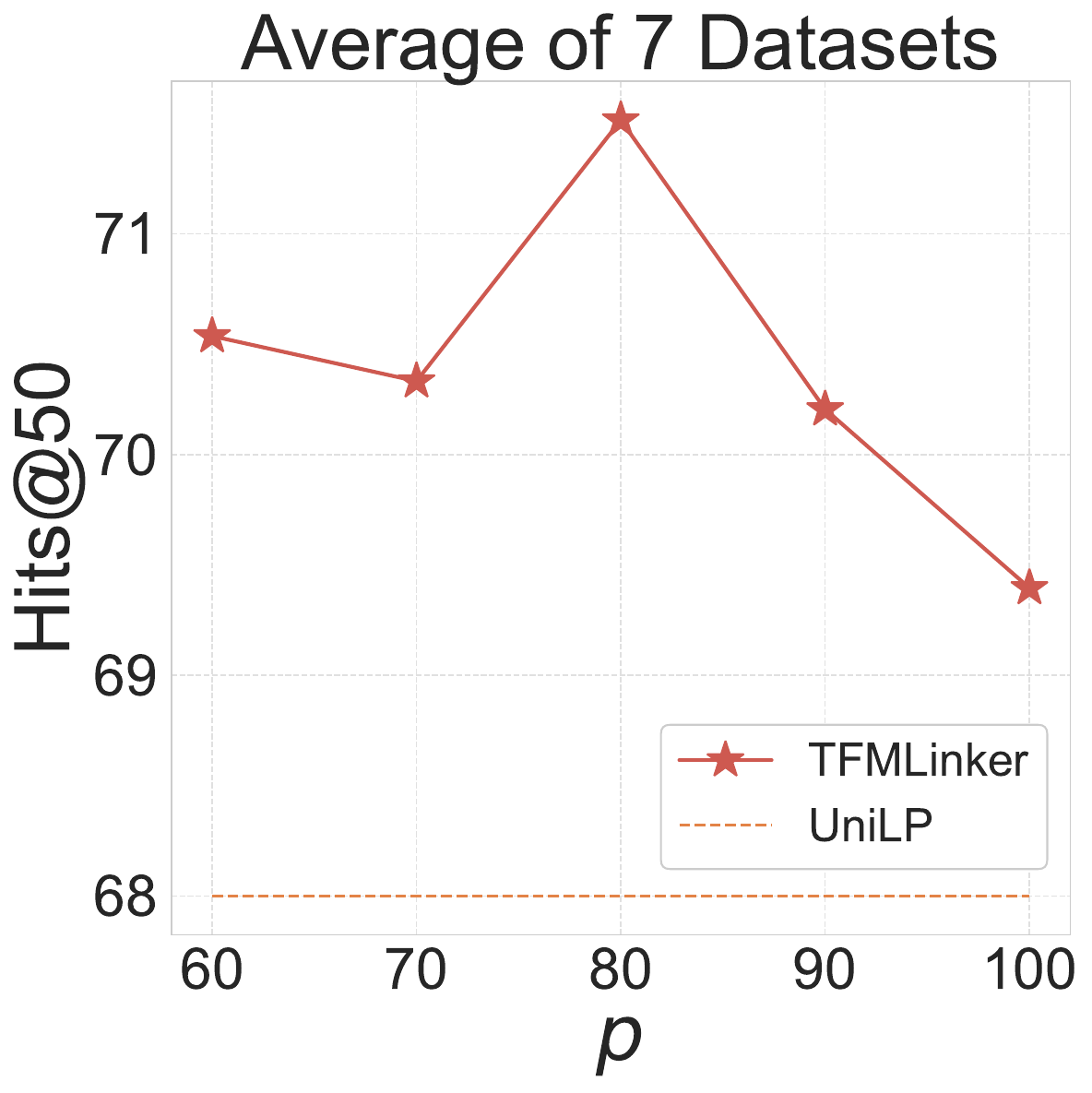}
	\end{subfigure}
    \vspace{-0.3cm}
    \caption{The impact of the number of learnable prototypes $p$.}
    \label{fig:hp-sensitivity}
    \vspace{0.3cm}
\end{figure*}

To answer \textbf{RQ3}, we evaluate three key modules of \name: the \context~module, the \encoder, and the TFM. For the the first module, we compare against a variant that removes the global context, retaining only the local context (denoted as \textit{w/o global context}). For the encoder, we test three configurations:
\begin{itemize}[leftmargin=0.5cm]
    \item w/o GNN pre-train: we use random initialization instead of pre-trained GNN weights.
    \item w/o GNN module: we remove the GNN module.
    \item w/o heuristics: we exclude of heuristic methods.
\end{itemize}
For the TFM, we examine the effect of random initialization in place of pre-trained TFM weights (denoted as w/o TFM pre-train). The results are shown in Table~\ref{tab:ablation}. We have the following observations:

When the global context is removed, our method yields suboptimal performance, indicating that the global context plays a crucial role in the in-context learning. Without the global context, our module essentially reduces to random sampling, losing its ability to capture common and transferable patterns that are shared across diverse tasks and graphs.

The ablation experiments on the encoder similarly lead to performance degradation. Replacing pre-trained GNN weights with random initialization results in noticeable performance drop, which underscores the importance of a pre-trained encoder in enabling the TFM to grasp topological graph structures. One plausible explanation is that without pre-training, the GNN tends to overfit to the specific contextual relationships present in a particular graph dataset, thereby losing its ability to capture fine-grained topological information. Moreover, removing either the entire GNN module or heuristic methods consistently degrades the results, confirming that diverse link encoders are crucial for capturing rich structural information in graphs.

Replacing pre-trained TFM weights with random initialization leads to a noticeable decline in performance, highlighting the importance of using a pre-trained TFM for effective in-context learning across diverse link prediction benchmarks. Considering that the TFMs are pre-trained on synthetic data generated under causal models, these results also confirm that pre-training on such synthetic data contributes to better adaptation to downstream scenarios.

\subsubsection{Hyper-parameter Sensitivity}

We further evaluate the sensitivity of the number of learnable prototypes $p$, with UniLP employed as the base model. The results on USAir, Cora, Facebook, and average performance are shown in Figure~\ref{fig:hp-sensitivity}. This parameter plays a crucial role in balancing the modeling scope: if set too high, the model may overemphasize common and transferable patterns across diverse tasks and graphs, whereas if set too low, it risks learning only patterns specific to a single task. While an appropriate choice of $p$ can enhance performance, our method is not highly sensitive to this hyperparameter and consistently outperforms the best baselines across a wide range of values.

\section{Related Work}\label{sec:related}

\textbf{Link Prediction.} The field of link prediction has long been dominated by two primary paradigms: heuristic methods and GNNs for Link Prediction (GNN4LP). Heuristic approaches rely on manually designed topological measures, such as local similarity indices~\cite{adamic2003friends,barabasi1999emergence,liben2003link,zhou2009predicting} or global path-based metrics~\cite{katz1953new,page1999pagerank}, to estimate the likelihood of missing edges. On the other hand, GNN4LP methods learn structural representations through data-driven architectures. GAE~\cite{kipf2016variational} pioneered the use of graph autoencoders to learn node embeddings for link scoring. SEAL~\cite{zhang2018link} further argued that effective link prediction requires explicit link-level structural encoding and introduced the labeling trick to enable GNNs to capture pairwise structural patterns. Subsequent works such as ELPH~\cite{chamberlaingraph}, NCNC~\cite{wangneural}, and MPLP~\cite{dong2024pure} have pushed scalability and performance, establishing strong results across standard benchmarks. Nevertheless, both paradigms face limitations: heuristic methods cannot fully capture effective structural features, while GNN4LP methods lack generalization across datasets or domains. These shortcomings constrain their practical applicability in real-world scenarios.

\textbf{Graph Foundation Model.} GFMs have attracted considerable attention in graph machine learning. The main goal is to facilitate effective knowledge transfer across diverse graph datasets, enabling models to learn generalizable knowledge from various graph tasks and apply it successfully to unseen graph structures. Architecturally, GFMs are organized into three primary categories: GNN-based~\cite{huang2023prodigy,song2025scalable,dong2025universal}, LLM-based~\cite{chen2024exploring}, and GNN+LLM-based models~\cite{liuone,yang2025harnessing}. GNN-based models capture topological patterns through graph-structured inductive biases. For instance, UniLP~\cite{dong2025universal} employs in-context learning driven by GNN and attention mechanisms to generalize across unseen graph datasets without requiring task-specific training. LLM-based models integrate semantic understanding via natural language processing. As an example, InstructGLM~\cite{ye2024language} expands the vocabulary of a pre-trained language model by representing node features as distinct tokens, thereby enhancing cross-modal graph comprehension. GNN+LLM-based models combine structural and semantic information to produce more generalized representations. This is exemplified by OFA~\cite{liuone}, which aligns textual descriptions of nodes and edges into a unified embedding space through language modeling, enabling cross-modal graph generalization. Nevertheless, existing methods still face inherent limitations due to their foundational architectures, such as limited pre-training scale~\cite{liu2025graph} or heavy reliance on textual information~\cite{eremeev2025turning,eremeev2025graphpfn}.

\textbf{Tabular Foundation Model.} Recently, TabPFN~\cite{hollmanntabpfn,hollmann2025accurate} and LimiX~\cite{zhang2025limix} have demonstrated state-of-the-art performance and efficiency on small-to-medium-scale tabular data by fitting diverse generative processes with prior data, contributing to the growing recognition of TFMs as a distinct paradigm. Concurrent works such as TuneTables~\cite{feuer2024tunetables} and BETA~\cite{liutabpfn} further scale these approaches to real-world and larger datasets, narrowing the performance gap between TFMs and classical methods in scenarios with larger sample sizes. Currently, new TFMs are emerging regularly, and their success has extended beyond pure tabular tasks to applications such as time series forecasting~\cite{hoo2025tables}, node classification~\cite{eremeev2025turning,eremeev2025graphpfn}, and graph anomaly detection~\cite{liu2026tabularfoundationmodelsstrong}. In this work, we contribute to this expanding scope by demonstrating, for the first time, that TFMs can also effectively serve as a powerful core component for the challenging task of universal link prediction across diverse graphs, which further broadens the potential application landscape of TFMs.

\section{Conclusion}\label{sec:con}
In this work, we propose \name, a novel link prediction approach across graph datasets and domains, leveraging TFMs for flexible adaptation to new graphs without fine-tuning. Our method effectively achieves this goal by integrating three key components, including the \context~module, the \encoder, and the TFM. Extensive experiments on 6 diverse graph benchmark datasets demonstrate the superiority of \name~over state-of-the-art methods without requiring dataset-specific finetuning.

\bibliographystyle{ACM-Reference-Format}
\balance
\bibliography{base}


\begin{thebibliography}{52}


\ifx \showCODEN    \undefined \def \showCODEN     #1{\unskip}     \fi
\ifx \showISBNx    \undefined \def \showISBNx     #1{\unskip}     \fi
\ifx \showISBNxiii \undefined \def \showISBNxiii  #1{\unskip}     \fi
\ifx \showISSN     \undefined \def \showISSN      #1{\unskip}     \fi
\ifx \showLCCN     \undefined \def \showLCCN      #1{\unskip}     \fi
\ifx \shownote     \undefined \def \shownote      #1{#1}          \fi
\ifx \showarticletitle \undefined \def \showarticletitle #1{#1}   \fi
\ifx \showURL      \undefined \def \showURL       {\relax}        \fi
\providecommand\bibfield[2]{#2}
\providecommand\bibinfo[2]{#2}
\providecommand\natexlab[1]{#1}
\providecommand\showeprint[2][]{arXiv:#2}

\bibitem[new(2006)]%
        {newman2006finding}
 \bibinfo{year}{2006}\natexlab{}.
\newblock \showarticletitle{Finding community structure in networks using the eigenvectors of matrices}.
\newblock \bibinfo{journal}{\emph{Physical Review E—Statistical, Nonlinear, and Soft Matter Physics}} \bibinfo{volume}{74}, \bibinfo{number}{3} (\bibinfo{year}{2006}), \bibinfo{pages}{036104}.
\newblock


\bibitem[Ackland(2005)]%
        {ackland2005mapping}
\bibfield{author}{\bibinfo{person}{Robert Ackland}.} \bibinfo{year}{2005}\natexlab{}.
\newblock \showarticletitle{Mapping the US political blogosphere: Are conservative bloggers more prominent?} BlogTalk Downunder 2005 Conference, Sydney.
\newblock


\bibitem[Adamic and Adar(2003)]%
        {adamic2003friends}
\bibfield{author}{\bibinfo{person}{Lada~A Adamic} {and} \bibinfo{person}{Eytan Adar}.} \bibinfo{year}{2003}\natexlab{}.
\newblock \showarticletitle{Friends and neighbors on the web}.
\newblock \bibinfo{journal}{\emph{Social networks}} \bibinfo{volume}{25}, \bibinfo{number}{3} (\bibinfo{year}{2003}), \bibinfo{pages}{211--230}.
\newblock


\bibitem[Adamic and Glance(2005)]%
        {adamic2005political}
\bibfield{author}{\bibinfo{person}{Lada~A Adamic} {and} \bibinfo{person}{Natalie Glance}.} \bibinfo{year}{2005}\natexlab{}.
\newblock \showarticletitle{The political blogosphere and the 2004 US election: divided they blog}. In \bibinfo{booktitle}{\emph{Proceedings of the 3rd international workshop on Link discovery}}. \bibinfo{pages}{36--43}.
\newblock


\bibitem[Barab{\'a}si and Albert(1999)]%
        {barabasi1999emergence}
\bibfield{author}{\bibinfo{person}{Albert-L{\'a}szl{\'o} Barab{\'a}si} {and} \bibinfo{person}{R{\'e}ka Albert}.} \bibinfo{year}{1999}\natexlab{}.
\newblock \showarticletitle{Emergence of scaling in random networks}.
\newblock \bibinfo{journal}{\emph{science}} \bibinfo{volume}{286}, \bibinfo{number}{5439} (\bibinfo{year}{1999}), \bibinfo{pages}{509--512}.
\newblock


\bibitem[Chamberlain et~al\mbox{.}({[n.\,d.]})]%
        {chamberlaingraph}
\bibfield{author}{\bibinfo{person}{Benjamin~Paul Chamberlain}, \bibinfo{person}{Sergey Shirobokov}, \bibinfo{person}{Emanuele Rossi}, \bibinfo{person}{Fabrizio Frasca}, \bibinfo{person}{Thomas Markovich}, \bibinfo{person}{Nils~Yannick Hammerla}, \bibinfo{person}{Michael~M Bronstein}, {and} \bibinfo{person}{Max Hansmire}.} \bibinfo{year}{[n.\,d.]}\natexlab{}.
\newblock \showarticletitle{Graph Neural Networks for Link Prediction with Subgraph Sketching}. In \bibinfo{booktitle}{\emph{The Eleventh International Conference on Learning Representations}}.
\newblock


\bibitem[Chen et~al\mbox{.}(2024)]%
        {chen2024exploring}
\bibfield{author}{\bibinfo{person}{Zhikai Chen}, \bibinfo{person}{Haitao Mao}, \bibinfo{person}{Hang Li}, \bibinfo{person}{Wei Jin}, \bibinfo{person}{Hongzhi Wen}, \bibinfo{person}{Xiaochi Wei}, \bibinfo{person}{Shuaiqiang Wang}, \bibinfo{person}{Dawei Yin}, \bibinfo{person}{Wenqi Fan}, \bibinfo{person}{Hui Liu}, {et~al\mbox{.}}} \bibinfo{year}{2024}\natexlab{}.
\newblock \showarticletitle{Exploring the potential of large language models (llms) in learning on graphs}.
\newblock \bibinfo{journal}{\emph{ACM SIGKDD Explorations Newsletter}} \bibinfo{volume}{25}, \bibinfo{number}{2} (\bibinfo{year}{2024}), \bibinfo{pages}{42--61}.
\newblock


\bibitem[Daud et~al\mbox{.}(2020)]%
        {daud2020applications}
\bibfield{author}{\bibinfo{person}{Nur~Nasuha Daud}, \bibinfo{person}{Siti~Hafizah Ab~Hamid}, \bibinfo{person}{Muntadher Saadoon}, \bibinfo{person}{Firdaus Sahran}, {and} \bibinfo{person}{Nor~Badrul Anuar}.} \bibinfo{year}{2020}\natexlab{}.
\newblock \showarticletitle{Applications of link prediction in social networks: A review}.
\newblock \bibinfo{journal}{\emph{Journal of Network and Computer Applications}}  \bibinfo{volume}{166} (\bibinfo{year}{2020}), \bibinfo{pages}{102716}.
\newblock


\bibitem[Dong et~al\mbox{.}(2024)]%
        {dong2024pure}
\bibfield{author}{\bibinfo{person}{Kaiwen Dong}, \bibinfo{person}{Zhichun Guo}, {and} \bibinfo{person}{Nitesh Chawla}.} \bibinfo{year}{2024}\natexlab{}.
\newblock \showarticletitle{Pure message passing can estimate common neighbor for link prediction}.
\newblock \bibinfo{journal}{\emph{Advances in Neural Information Processing Systems}}  \bibinfo{volume}{37} (\bibinfo{year}{2024}), \bibinfo{pages}{73000--73035}.
\newblock


\bibitem[Dong et~al\mbox{.}(2025)]%
        {dong2025universal}
\bibfield{author}{\bibinfo{person}{Kaiwen Dong}, \bibinfo{person}{Haitao Mao}, \bibinfo{person}{Zhichun Guo}, {and} \bibinfo{person}{Nitesh~V Chawla}.} \bibinfo{year}{2025}\natexlab{}.
\newblock \showarticletitle{Universal Link Predictor By In-Context Learning on Graphs}.
\newblock \bibinfo{journal}{\emph{Transactions on Machine Learning Research}} (\bibinfo{year}{2025}).
\newblock
\showISSN{2835-8856}


\bibitem[Eremeev et~al\mbox{.}(2025a)]%
        {eremeev2025turning}
\bibfield{author}{\bibinfo{person}{Dmitry Eremeev}, \bibinfo{person}{Gleb Bazhenov}, \bibinfo{person}{Oleg Platonov}, \bibinfo{person}{Artem Babenko}, {and} \bibinfo{person}{Liudmila Prokhorenkova}.} \bibinfo{year}{2025}\natexlab{a}.
\newblock \showarticletitle{Turning Tabular Foundation Models into Graph Foundation Models}. In \bibinfo{booktitle}{\emph{New Perspectives in Graph Machine Learning}}.
\newblock


\bibitem[Eremeev et~al\mbox{.}(2025b)]%
        {eremeev2025graphpfn}
\bibfield{author}{\bibinfo{person}{Dmitry Eremeev}, \bibinfo{person}{Oleg Platonov}, \bibinfo{person}{Gleb Bazhenov}, \bibinfo{person}{Artem Babenko}, {and} \bibinfo{person}{Liudmila Prokhorenkova}.} \bibinfo{year}{2025}\natexlab{b}.
\newblock \showarticletitle{GraphPFN: A prior-data fitted graph foundation model}.
\newblock \bibinfo{journal}{\emph{arXiv preprint arXiv:2509.21489}} (\bibinfo{year}{2025}).
\newblock


\bibitem[Feuer et~al\mbox{.}(2024)]%
        {feuer2024tunetables}
\bibfield{author}{\bibinfo{person}{Benjamin Feuer}, \bibinfo{person}{Robin~T Schirrmeister}, \bibinfo{person}{Valeriia Cherepanova}, \bibinfo{person}{Chinmay Hegde}, \bibinfo{person}{Frank Hutter}, \bibinfo{person}{Micah Goldblum}, \bibinfo{person}{Niv Cohen}, {and} \bibinfo{person}{Colin White}.} \bibinfo{year}{2024}\natexlab{}.
\newblock \showarticletitle{Tunetables: Context optimization for scalable prior-data fitted networks}.
\newblock \bibinfo{journal}{\emph{Advances in Neural Information Processing Systems}}  \bibinfo{volume}{37} (\bibinfo{year}{2024}), \bibinfo{pages}{83430--83464}.
\newblock


\bibitem[Hamilton et~al\mbox{.}(2017)]%
        {hamilton2017inductive}
\bibfield{author}{\bibinfo{person}{Will Hamilton}, \bibinfo{person}{Zhitao Ying}, {and} \bibinfo{person}{Jure Leskovec}.} \bibinfo{year}{2017}\natexlab{}.
\newblock \showarticletitle{Inductive representation learning on large graphs}.
\newblock \bibinfo{journal}{\emph{Advances in neural information processing systems}}  \bibinfo{volume}{30} (\bibinfo{year}{2017}).
\newblock


\bibitem[Hollmann et~al\mbox{.}({[n.\,d.]})]%
        {hollmanntabpfn}
\bibfield{author}{\bibinfo{person}{Noah Hollmann}, \bibinfo{person}{Samuel M{\"u}ller}, \bibinfo{person}{Katharina Eggensperger}, {and} \bibinfo{person}{Frank Hutter}.} \bibinfo{year}{[n.\,d.]}\natexlab{}.
\newblock \showarticletitle{TabPFN: A Transformer That Solves Small Tabular Classification Problems in a Second}. In \bibinfo{booktitle}{\emph{The Eleventh International Conference on Learning Representations}}.
\newblock


\bibitem[Hollmann et~al\mbox{.}(2025)]%
        {hollmann2025accurate}
\bibfield{author}{\bibinfo{person}{Noah Hollmann}, \bibinfo{person}{Samuel M{\"u}ller}, \bibinfo{person}{Lennart Purucker}, \bibinfo{person}{Arjun Krishnakumar}, \bibinfo{person}{Max K{\"o}rfer}, \bibinfo{person}{Shi~Bin Hoo}, \bibinfo{person}{Robin~Tibor Schirrmeister}, {and} \bibinfo{person}{Frank Hutter}.} \bibinfo{year}{2025}\natexlab{}.
\newblock \showarticletitle{Accurate predictions on small data with a tabular foundation model}.
\newblock \bibinfo{journal}{\emph{Nature}} \bibinfo{volume}{637}, \bibinfo{number}{8045} (\bibinfo{year}{2025}), \bibinfo{pages}{319--326}.
\newblock


\bibitem[Hoo et~al\mbox{.}(2025)]%
        {hoo2025tables}
\bibfield{author}{\bibinfo{person}{Shi~Bin Hoo}, \bibinfo{person}{Samuel M{\"u}ller}, \bibinfo{person}{David Salinas}, {and} \bibinfo{person}{Frank Hutter}.} \bibinfo{year}{2025}\natexlab{}.
\newblock \showarticletitle{From Tables to Time: How TabPFN-v2 Outperforms Specialized Time Series Forecasting Models}.
\newblock \bibinfo{journal}{\emph{arXiv preprint arXiv:2501.02945}} (\bibinfo{year}{2025}).
\newblock


\bibitem[Huang et~al\mbox{.}(2023)]%
        {huang2023prodigy}
\bibfield{author}{\bibinfo{person}{Qian Huang}, \bibinfo{person}{Hongyu Ren}, \bibinfo{person}{Peng Chen}, \bibinfo{person}{Gregor Kr{\v{z}}manc}, \bibinfo{person}{Daniel Zeng}, \bibinfo{person}{Percy~S Liang}, {and} \bibinfo{person}{Jure Leskovec}.} \bibinfo{year}{2023}\natexlab{}.
\newblock \showarticletitle{Prodigy: Enabling in-context learning over graphs}.
\newblock \bibinfo{journal}{\emph{Advances in Neural Information Processing Systems}}  \bibinfo{volume}{36} (\bibinfo{year}{2023}), \bibinfo{pages}{16302--16317}.
\newblock


\bibitem[Katz(1953)]%
        {katz1953new}
\bibfield{author}{\bibinfo{person}{Leo Katz}.} \bibinfo{year}{1953}\natexlab{}.
\newblock \showarticletitle{A new status index derived from sociometric analysis}.
\newblock \bibinfo{journal}{\emph{Psychometrika}} \bibinfo{volume}{18}, \bibinfo{number}{1} (\bibinfo{year}{1953}), \bibinfo{pages}{39--43}.
\newblock


\bibitem[Kipf and Welling(2016)]%
        {kipf2016variational}
\bibfield{author}{\bibinfo{person}{Thomas~N Kipf} {and} \bibinfo{person}{Max Welling}.} \bibinfo{year}{2016}\natexlab{}.
\newblock \showarticletitle{Variational Graph Auto-Encoders}.
\newblock \bibinfo{journal}{\emph{NIPS Workshop on Bayesian Deep Learning}} (\bibinfo{year}{2016}).
\newblock


\bibitem[Kov{\'a}cs et~al\mbox{.}(2019)]%
        {kovacs2019network}
\bibfield{author}{\bibinfo{person}{Istv{\'a}n~A Kov{\'a}cs}, \bibinfo{person}{Katja Luck}, \bibinfo{person}{Kerstin Spirohn}, \bibinfo{person}{Yang Wang}, \bibinfo{person}{Carl Pollis}, \bibinfo{person}{Sadie Schlabach}, \bibinfo{person}{Wenting Bian}, \bibinfo{person}{Dae-Kyum Kim}, \bibinfo{person}{Nishka Kishore}, \bibinfo{person}{Tong Hao}, {et~al\mbox{.}}} \bibinfo{year}{2019}\natexlab{}.
\newblock \showarticletitle{Network-based prediction of protein interactions}.
\newblock \bibinfo{journal}{\emph{Nature communications}} \bibinfo{volume}{10}, \bibinfo{number}{1} (\bibinfo{year}{2019}), \bibinfo{pages}{1240}.
\newblock


\bibitem[Li et~al\mbox{.}(2023)]%
        {li2023evaluating}
\bibfield{author}{\bibinfo{person}{Juanhui Li}, \bibinfo{person}{Harry Shomer}, \bibinfo{person}{Haitao Mao}, \bibinfo{person}{Shenglai Zeng}, \bibinfo{person}{Yao Ma}, \bibinfo{person}{Neil Shah}, \bibinfo{person}{Jiliang Tang}, {and} \bibinfo{person}{Dawei Yin}.} \bibinfo{year}{2023}\natexlab{}.
\newblock \showarticletitle{Evaluating graph neural networks for link prediction: Current pitfalls and new benchmarking}.
\newblock \bibinfo{journal}{\emph{Advances in Neural Information Processing Systems}}  \bibinfo{volume}{36} (\bibinfo{year}{2023}), \bibinfo{pages}{3853--3866}.
\newblock


\bibitem[Li et~al\mbox{.}(2020)]%
        {li2020distance}
\bibfield{author}{\bibinfo{person}{Pan Li}, \bibinfo{person}{Yanbang Wang}, \bibinfo{person}{Hongwei Wang}, {and} \bibinfo{person}{Jure Leskovec}.} \bibinfo{year}{2020}\natexlab{}.
\newblock \showarticletitle{Distance encoding: Design provably more powerful neural networks for graph representation learning}.
\newblock \bibinfo{journal}{\emph{Advances in Neural Information Processing Systems}}  \bibinfo{volume}{33} (\bibinfo{year}{2020}), \bibinfo{pages}{4465--4478}.
\newblock


\bibitem[Liben-Nowell and Kleinberg(2003)]%
        {liben2003link}
\bibfield{author}{\bibinfo{person}{David Liben-Nowell} {and} \bibinfo{person}{Jon Kleinberg}.} \bibinfo{year}{2003}\natexlab{}.
\newblock \showarticletitle{The link prediction problem for social networks}. In \bibinfo{booktitle}{\emph{Proceedings of the twelfth international conference on Information and knowledge management}}. \bibinfo{pages}{556--559}.
\newblock


\bibitem[Lin et~al\mbox{.}(2022)]%
        {lin2022prototypical}
\bibfield{author}{\bibinfo{person}{Shuai Lin}, \bibinfo{person}{Chen Liu}, \bibinfo{person}{Pan Zhou}, \bibinfo{person}{Zi-Yuan Hu}, \bibinfo{person}{Shuojia Wang}, \bibinfo{person}{Ruihui Zhao}, \bibinfo{person}{Yefeng Zheng}, \bibinfo{person}{Liang Lin}, \bibinfo{person}{Eric Xing}, {and} \bibinfo{person}{Xiaodan Liang}.} \bibinfo{year}{2022}\natexlab{}.
\newblock \showarticletitle{Prototypical graph contrastive learning}.
\newblock \bibinfo{journal}{\emph{IEEE transactions on neural networks and learning systems}} \bibinfo{volume}{35}, \bibinfo{number}{2} (\bibinfo{year}{2022}), \bibinfo{pages}{2747--2758}.
\newblock


\bibitem[Liu et~al\mbox{.}({[n.\,d.]})]%
        {liuone}
\bibfield{author}{\bibinfo{person}{Hao Liu}, \bibinfo{person}{Jiarui Feng}, \bibinfo{person}{Lecheng Kong}, \bibinfo{person}{Ningyue Liang}, \bibinfo{person}{Dacheng Tao}, \bibinfo{person}{Yixin Chen}, {and} \bibinfo{person}{Muhan Zhang}.} \bibinfo{year}{[n.\,d.]}\natexlab{}.
\newblock \showarticletitle{One For All: Towards Training One Graph Model For All Classification Tasks}. In \bibinfo{booktitle}{\emph{The Twelfth International Conference on Learning Representations}}.
\newblock


\bibitem[Liu et~al\mbox{.}(2025)]%
        {liu2025graph}
\bibfield{author}{\bibinfo{person}{Jiawei Liu}, \bibinfo{person}{Cheng Yang}, \bibinfo{person}{Zhiyuan Lu}, \bibinfo{person}{Junze Chen}, \bibinfo{person}{Yibo Li}, \bibinfo{person}{Mengmei Zhang}, \bibinfo{person}{Ting Bai}, \bibinfo{person}{Yuan Fang}, \bibinfo{person}{Lichao Sun}, \bibinfo{person}{Philip~S Yu}, {et~al\mbox{.}}} \bibinfo{year}{2025}\natexlab{}.
\newblock \showarticletitle{Graph foundation models: Concepts, opportunities and challenges}.
\newblock \bibinfo{journal}{\emph{IEEE Transactions on Pattern Analysis and Machine Intelligence}} (\bibinfo{year}{2025}).
\newblock


\bibitem[Liu and Ye({[n.\,d.]})]%
        {liutabpfn}
\bibfield{author}{\bibinfo{person}{Siyang Liu} {and} \bibinfo{person}{Han-Jia Ye}.} \bibinfo{year}{[n.\,d.]}\natexlab{}.
\newblock \showarticletitle{TabPFN Unleashed: A Scalable and Effective Solution to Tabular Classification Problems}. In \bibinfo{booktitle}{\emph{Forty-second International Conference on Machine Learning}}.
\newblock


\bibitem[Liu et~al\mbox{.}(2026)]%
        {liu2026tabularfoundationmodelsstrong}
\bibfield{author}{\bibinfo{person}{Yunhui Liu}, \bibinfo{person}{Tieke He}, \bibinfo{person}{Yongchao Liu}, \bibinfo{person}{Can Yi}, \bibinfo{person}{Hong Jin}, {and} \bibinfo{person}{Chuntao Hong}.} \bibinfo{year}{2026}\natexlab{}.
\newblock \showarticletitle{Tabular Foundation Models are Strong Graph Anomaly Detectors}.
\newblock \bibinfo{journal}{\emph{arXiv preprint arXiv:2601.17301}} (\bibinfo{year}{2026}).
\newblock


\bibitem[Ma et~al\mbox{.}(2024b)]%
        {ma2024context}
\bibfield{author}{\bibinfo{person}{Junwei Ma}, \bibinfo{person}{Valentin Thomas}, \bibinfo{person}{Guangwei Yu}, {and} \bibinfo{person}{Anthony Caterini}.} \bibinfo{year}{2024}\natexlab{b}.
\newblock \showarticletitle{In-context data distillation with TabPFN}.
\newblock \bibinfo{journal}{\emph{arXiv preprint arXiv:2402.06971}} (\bibinfo{year}{2024}).
\newblock


\bibitem[Ma et~al\mbox{.}(2024a)]%
        {ma2024mixture}
\bibfield{author}{\bibinfo{person}{Li Ma}, \bibinfo{person}{Haoyu Han}, \bibinfo{person}{Juanhui Li}, \bibinfo{person}{Harry Shomer}, \bibinfo{person}{Hui Liu}, \bibinfo{person}{Xiaofeng Gao}, {and} \bibinfo{person}{Jiliang Tang}.} \bibinfo{year}{2024}\natexlab{a}.
\newblock \showarticletitle{Mixture of link predictors on graphs}.
\newblock \bibinfo{journal}{\emph{Advances in Neural Information Processing Systems}}  \bibinfo{volume}{37} (\bibinfo{year}{2024}), \bibinfo{pages}{16043--16070}.
\newblock


\bibitem[Mao et~al\mbox{.}({[n.\,d.]})]%
        {maorevisiting}
\bibfield{author}{\bibinfo{person}{Haitao Mao}, \bibinfo{person}{Juanhui Li}, \bibinfo{person}{Harry Shomer}, \bibinfo{person}{Bingheng Li}, \bibinfo{person}{Wenqi Fan}, \bibinfo{person}{Yao Ma}, \bibinfo{person}{Tong Zhao}, \bibinfo{person}{Neil Shah}, {and} \bibinfo{person}{Jiliang Tang}.} \bibinfo{year}{[n.\,d.]}\natexlab{}.
\newblock \showarticletitle{Revisiting Link Prediction: a data perspective}. In \bibinfo{booktitle}{\emph{The Twelfth International Conference on Learning Representations}}.
\newblock


\bibitem[Page et~al\mbox{.}(1999)]%
        {page1999pagerank}
\bibfield{author}{\bibinfo{person}{Lawrence Page}, \bibinfo{person}{Sergey Brin}, \bibinfo{person}{Rajeev Motwani}, {and} \bibinfo{person}{Terry Winograd}.} \bibinfo{year}{1999}\natexlab{}.
\newblock \bibinfo{booktitle}{\emph{The PageRank citation ranking: Bringing order to the web.}}
\newblock \bibinfo{type}{{T}echnical {R}eport}. \bibinfo{institution}{Stanford infolab}.
\newblock


\bibitem[Rozemberczki et~al\mbox{.}(2021)]%
        {rozemberczki2021multi}
\bibfield{author}{\bibinfo{person}{Benedek Rozemberczki}, \bibinfo{person}{Carl Allen}, {and} \bibinfo{person}{Rik Sarkar}.} \bibinfo{year}{2021}\natexlab{}.
\newblock \showarticletitle{Multi-scale attributed node embedding}.
\newblock \bibinfo{journal}{\emph{Journal of Complex Networks}} \bibinfo{volume}{9}, \bibinfo{number}{2} (\bibinfo{year}{2021}), \bibinfo{pages}{cnab014}.
\newblock


\bibitem[Rubachev et~al\mbox{.}(2025)]%
        {rubachev2025finetuningtabularfoundationmodels}
\bibfield{author}{\bibinfo{person}{Ivan Rubachev}, \bibinfo{person}{Akim Kotelnikov}, \bibinfo{person}{Nikolay Kartashev}, {and} \bibinfo{person}{Artem Babenko}.} \bibinfo{year}{2025}\natexlab{}.
\newblock \showarticletitle{On finetuning tabular foundation models}.
\newblock \bibinfo{journal}{\emph{arXiv preprint arXiv:2506.08982}} (\bibinfo{year}{2025}).
\newblock


\bibitem[Shchur et~al\mbox{.}(2018)]%
        {shchur2018pitfalls}
\bibfield{author}{\bibinfo{person}{Oleksandr Shchur}, \bibinfo{person}{Maximilian Mumme}, \bibinfo{person}{Aleksandar Bojchevski}, {and} \bibinfo{person}{Stephan G{\"u}nnemann}.} \bibinfo{year}{2018}\natexlab{}.
\newblock \showarticletitle{Pitfalls of graph neural network evaluation}.
\newblock \bibinfo{journal}{\emph{arXiv preprint arXiv:1811.05868}} (\bibinfo{year}{2018}).
\newblock


\bibitem[Shomer et~al\mbox{.}(2024)]%
        {shomer2024lpformer}
\bibfield{author}{\bibinfo{person}{Harry Shomer}, \bibinfo{person}{Yao Ma}, \bibinfo{person}{Haitao Mao}, \bibinfo{person}{Juanhui Li}, \bibinfo{person}{Bo Wu}, {and} \bibinfo{person}{Jiliang Tang}.} \bibinfo{year}{2024}\natexlab{}.
\newblock \showarticletitle{Lpformer: An adaptive graph transformer for link prediction}. In \bibinfo{booktitle}{\emph{Proceedings of the 30th ACM SIGKDD conference on knowledge discovery and data mining}}. \bibinfo{pages}{2686--2698}.
\newblock


\bibitem[Song et~al\mbox{.}(2025)]%
        {song2025scalable}
\bibfield{author}{\bibinfo{person}{Yu Song}, \bibinfo{person}{Zhigang Hua}, \bibinfo{person}{Harry Shomer}, \bibinfo{person}{Yan Xie}, \bibinfo{person}{Jingzhe Liu}, \bibinfo{person}{Bo Long}, {and} \bibinfo{person}{Hui Liu}.} \bibinfo{year}{2025}\natexlab{}.
\newblock \showarticletitle{A Scalable Pretraining Framework for Link Prediction with Efficient Adaptation}. In \bibinfo{booktitle}{\emph{Proceedings of the 31st ACM SIGKDD Conference on Knowledge Discovery and Data Mining V. 2}}. \bibinfo{pages}{2621--2632}.
\newblock


\bibitem[Spring et~al\mbox{.}(2002)]%
        {spring2002measuring}
\bibfield{author}{\bibinfo{person}{Neil Spring}, \bibinfo{person}{Ratul Mahajan}, {and} \bibinfo{person}{David Wetherall}.} \bibinfo{year}{2002}\natexlab{}.
\newblock \showarticletitle{Measuring ISP topologies with Rocketfuel}.
\newblock \bibinfo{journal}{\emph{ACM SIGCOMM Computer Communication Review}} \bibinfo{volume}{32}, \bibinfo{number}{4} (\bibinfo{year}{2002}), \bibinfo{pages}{133--145}.
\newblock


\bibitem[Von~Mering et~al\mbox{.}(2002)]%
        {von2002comparative}
\bibfield{author}{\bibinfo{person}{Christian Von~Mering}, \bibinfo{person}{Roland Krause}, \bibinfo{person}{Berend Snel}, \bibinfo{person}{Michael Cornell}, \bibinfo{person}{Stephen~G Oliver}, \bibinfo{person}{Stanley Fields}, {and} \bibinfo{person}{Peer Bork}.} \bibinfo{year}{2002}\natexlab{}.
\newblock \showarticletitle{Comparative assessment of large-scale data sets of protein--protein interactions}.
\newblock \bibinfo{journal}{\emph{Nature}} \bibinfo{volume}{417}, \bibinfo{number}{6887} (\bibinfo{year}{2002}), \bibinfo{pages}{399--403}.
\newblock


\bibitem[Wang et~al\mbox{.}({[n.\,d.]})]%
        {wangneural}
\bibfield{author}{\bibinfo{person}{Xiyuan Wang}, \bibinfo{person}{Haotong Yang}, {and} \bibinfo{person}{Muhan Zhang}.} \bibinfo{year}{[n.\,d.]}\natexlab{}.
\newblock \showarticletitle{Neural Common Neighbor with Completion for Link Prediction}. In \bibinfo{booktitle}{\emph{The Twelfth International Conference on Learning Representations}}.
\newblock


\bibitem[Watts and Strogatz(1998)]%
        {watts1998collective}
\bibfield{author}{\bibinfo{person}{Duncan~J Watts} {and} \bibinfo{person}{Steven~H Strogatz}.} \bibinfo{year}{1998}\natexlab{}.
\newblock \showarticletitle{Collective dynamics of ‘small-world’networks}.
\newblock \bibinfo{journal}{\emph{nature}} \bibinfo{volume}{393}, \bibinfo{number}{6684} (\bibinfo{year}{1998}), \bibinfo{pages}{440--442}.
\newblock


\bibitem[Wu and Yin({[n.\,d.]})]%
        {wumoemeta}
\bibfield{author}{\bibinfo{person}{Han Wu} {and} \bibinfo{person}{Jie Yin}.} \bibinfo{year}{[n.\,d.]}\natexlab{}.
\newblock \showarticletitle{MoEMeta: Mixture-of-Experts Meta Learning for Few-Shot Relational Learning}. In \bibinfo{booktitle}{\emph{The Thirty-ninth Annual Conference on Neural Information Processing Systems}}.
\newblock


\bibitem[Yang et~al\mbox{.}(2025)]%
        {yang2025harnessing}
\bibfield{author}{\bibinfo{person}{Jinyu Yang}, \bibinfo{person}{Ruijia Wang}, \bibinfo{person}{Cheng Yang}, \bibinfo{person}{Bo Yan}, \bibinfo{person}{Qimin Zhou}, \bibinfo{person}{Yang Juan}, {and} \bibinfo{person}{Chuan Shi}.} \bibinfo{year}{2025}\natexlab{}.
\newblock \showarticletitle{Harnessing Language Model for Cross-Heterogeneity Graph Knowledge Transfer}. In \bibinfo{booktitle}{\emph{Proceedings of the AAAI Conference on Artificial Intelligence}}, Vol.~\bibinfo{volume}{39}. \bibinfo{pages}{13026--13034}.
\newblock


\bibitem[Yang et~al\mbox{.}(2016)]%
        {yang2016revisiting}
\bibfield{author}{\bibinfo{person}{Zhilin Yang}, \bibinfo{person}{William Cohen}, {and} \bibinfo{person}{Ruslan Salakhudinov}.} \bibinfo{year}{2016}\natexlab{}.
\newblock \showarticletitle{Revisiting semi-supervised learning with graph embeddings}. In \bibinfo{booktitle}{\emph{International conference on machine learning}}. PMLR, \bibinfo{pages}{40--48}.
\newblock


\bibitem[Ye et~al\mbox{.}(2025)]%
        {ye2025closer}
\bibfield{author}{\bibinfo{person}{Han-Jia Ye}, \bibinfo{person}{Si-Yang Liu}, {and} \bibinfo{person}{Wei-Lun Chao}.} \bibinfo{year}{2025}\natexlab{}.
\newblock \showarticletitle{A closer look at TabPFN v2: Understanding its strengths and extending its capabilities}. In \bibinfo{booktitle}{\emph{The Thirty-ninth Annual Conference on Neural Information Processing Systems}}.
\newblock


\bibitem[Ye et~al\mbox{.}(2024)]%
        {ye2024language}
\bibfield{author}{\bibinfo{person}{Ruosong Ye}, \bibinfo{person}{Caiqi Zhang}, \bibinfo{person}{Runhui Wang}, \bibinfo{person}{Shuyuan Xu}, {and} \bibinfo{person}{Yongfeng Zhang}.} \bibinfo{year}{2024}\natexlab{}.
\newblock \showarticletitle{Language is all a graph needs}. In \bibinfo{booktitle}{\emph{Findings of the association for computational linguistics: EACL 2024}}. \bibinfo{pages}{1955--1973}.
\newblock


\bibitem[Yun et~al\mbox{.}(2021)]%
        {yun2021neo}
\bibfield{author}{\bibinfo{person}{Seongjun Yun}, \bibinfo{person}{Seoyoon Kim}, \bibinfo{person}{Junhyun Lee}, \bibinfo{person}{Jaewoo Kang}, {and} \bibinfo{person}{Hyunwoo~J Kim}.} \bibinfo{year}{2021}\natexlab{}.
\newblock \showarticletitle{Neo-gnns: Neighborhood overlap-aware graph neural networks for link prediction}.
\newblock \bibinfo{journal}{\emph{Advances in Neural Information Processing Systems}}  \bibinfo{volume}{34} (\bibinfo{year}{2021}), \bibinfo{pages}{13683--13694}.
\newblock


\bibitem[Zhang and Chen(2018)]%
        {zhang2018link}
\bibfield{author}{\bibinfo{person}{Muhan Zhang} {and} \bibinfo{person}{Yixin Chen}.} \bibinfo{year}{2018}\natexlab{}.
\newblock \showarticletitle{Link prediction based on graph neural networks}.
\newblock \bibinfo{journal}{\emph{Advances in neural information processing systems}}  \bibinfo{volume}{31} (\bibinfo{year}{2018}).
\newblock


\bibitem[Zhang et~al\mbox{.}(2018)]%
        {zhang2018beyond}
\bibfield{author}{\bibinfo{person}{Muhan Zhang}, \bibinfo{person}{Zhicheng Cui}, \bibinfo{person}{Shali Jiang}, {and} \bibinfo{person}{Yixin Chen}.} \bibinfo{year}{2018}\natexlab{}.
\newblock \showarticletitle{Beyond link prediction: Predicting hyperlinks in adjacency space}. In \bibinfo{booktitle}{\emph{Proceedings of the AAAI conference on artificial intelligence}}, Vol.~\bibinfo{volume}{32}.
\newblock


\bibitem[Zhang et~al\mbox{.}(2025)]%
        {zhang2025limix}
\bibfield{author}{\bibinfo{person}{Xingxuan Zhang}, \bibinfo{person}{Gang Ren}, \bibinfo{person}{Han Yu}, \bibinfo{person}{Hao Yuan}, \bibinfo{person}{Hui Wang}, \bibinfo{person}{Jiansheng Li}, \bibinfo{person}{Jiayun Wu}, \bibinfo{person}{Lang Mo}, \bibinfo{person}{Li Mao}, \bibinfo{person}{Mingchao Hao}, {et~al\mbox{.}}} \bibinfo{year}{2025}\natexlab{}.
\newblock \showarticletitle{Limix: Unleashing structured-data modeling capability for generalist intelligence}.
\newblock \bibinfo{journal}{\emph{arXiv preprint arXiv:2509.03505}} (\bibinfo{year}{2025}).
\newblock


\bibitem[Zhou et~al\mbox{.}(2009)]%
        {zhou2009predicting}
\bibfield{author}{\bibinfo{person}{Tao Zhou}, \bibinfo{person}{Linyuan L{\"u}}, {and} \bibinfo{person}{Yi-Cheng Zhang}.} \bibinfo{year}{2009}\natexlab{}.
\newblock \showarticletitle{Predicting missing links via local information}.
\newblock \bibinfo{journal}{\emph{The European Physical Journal B}} \bibinfo{volume}{71}, \bibinfo{number}{4} (\bibinfo{year}{2009}), \bibinfo{pages}{623--630}.
\newblock


\end{thebibliography}

\appendix

\section{Experimental Details}

\begin{table*}
  \caption{Statistics of training datasets.}
  \begin{tabular}{lccccccc}
    \toprule
    Dataset&Domain&\# Nodes&\# Edges&Avg. node deg.&Std. node deg.&Max. node deg.&Density\\
    \midrule
    Ecoli&Biology&1,805&29,320&16.24&48.38&1,030&1.8009\%\\
    Yeast&Biology&2,375&23,386&9.85&15.5&118&0.8295\%\\
    Power&Transport&4,941&13,188&2.67&1.79&19&0.1081\%\\
    PolBlogs&Web&1,490&19,025&12.77&20.73&256&1.7150\%\\
    Router&Web&5,022&12,516&2.49&5.29&106&0.0993\%\\
    Physics&Collaboration&34,493&495,924&14.38&15.57&382&0.0834\%\\
    Pubmed&Citation&19,717&88,648&4.5&7.43&171&0.0456\%\\
    Citeseer&Citation&3,327&9,104&2.74&3.38&99&0.1645\%\\
    Twitch&Social&34,118&429,113&12.58&35.88&1,489&0.0737\%\\
    Github&Social&37,700&289,003&7.67&46.59&6,809&0.0407\%\\
    \bottomrule
  \end{tabular}
  \label{tab:train-dataset-detail}
\end{table*}

\begin{table*}
  \caption{Statistics of test datasets.}
  \begin{tabular}{lccccccc}
    \toprule
    Dataset&Domain&\# Nodes&\# Edges&Avg. node deg.&Std. node deg.&Max. node deg.&Density\\
    \midrule
    Celegans&Biology&297&4,296&14.46&12.97&134&9.7734\%\\
    USAir&Transport&332&4,252&12.81&20.13&139&7.7385\%\\
    PB&Web&1,222&33,428&27.36&38.42&351&4.4808\%\\
    NS&Collaboration&1,589&5,484&3.45&3.47&34&0.4347\%\\
    CS&Collaboration&18,333&163,788&8.93&9.11&136&0.0975\%\\
    Cora&Citation&2,708&10,556&3.9&5.23&168&0.2880\%\\
    Facebook&Social&22,470&171,002&7.61&15.26&472&0.0677\%\\
    \bottomrule
  \end{tabular}
  \label{tab:test-dataset-detail}
\end{table*}

\subsection{Datasets}\label{subsec:dataset-detail}

The complete details of both the training and test graph datasets are summarized in Table~\ref{tab:train-dataset-detail} and~\ref{tab:test-dataset-detail}. Following~\cite{dong2025universal}, these datasets are selected from diverse domains and exhibit varied graph statistics. They are intentionally chosen to ensure that \name~encounters a broad spectrum of link prediction connectivity patterns. This diversity in training data is essential for enabling \name~to adapt effectively to previously unseen graphs.

\subsection{Implementation Details}\label{subsec:implementation-detail}

We employ a 3-layer GraphSAGE ~\cite{hamilton2017inductive} with 128-dimensional hidden representations as the GNN and employ the LimiX-16M version as the default TFM. The heuristic-based link representations have a dimensionality of 16. The activation functions used are ReLU and GELU. The number of learnable prototypes is set to 80, and no dropout is applied. The model is trained for 5 epochs with a batch size of 16 and a learning rate of 1e-5. For each target link, we sample 40 positive and 40 negative links from the corresponding training dataset as in-context links. This design introduces variability: different target links from the same dataset, or the same target link across training batches, may be associated with different sets of in-context links. All experimental results are reported as mean ± standard deviation over 10 independent runs. The training phase of GFMs is guided by performance on a merged validation set comprising 200 links from the validation subset of each test dataset, while fine-tuning of the GNN relies solely on in-domain validation performance.

\section{Running Time Analysis}

To evaluate the computational overhead introduced by \name, we conduct a comparative analysis of average inference times between \name~and the in-context learning method UniLP on the Celegans dataset. All key factors influencing runtime—including batch size and hardware configuration (H200 GPU on a Linux server with 192-core Intel Xeon Platinum 8558 processors)—are kept constant during testing. UniLP requires 10.94 seconds, while \name~requires 12.02 seconds, demonstrating that \name~maintains computational efficiency comparable to the existing in-context learning approach. The slight additional overhead of \name~is attributed to the TFM, which has been demonstrated to deliver significant performance improvements. Compared to other baselines that require fine-tuning to adapt to downstream tasks, \name~incurs significantly lower computational costs as it eliminates the need for additional training on specific datasets.

\end{document}